\documentclass[12pt,twoside,reqno]{amsart}
\usepackage[colorlinks=true,citecolor=blue]{hyperref}
\usepackage{mathptmx, amsmath, amssymb, amsfonts, amsthm, mathptmx, enumerate, color}
\usepackage{graphicx}
\usepackage{epstopdf}

\theoremstyle{definition}

\numberwithin{equation}{section}
\usepackage{mathrsfs}

\usepackage{multirow}
\usepackage[ruled,vlined,linesnumbered]{algorithm2e}
\usepackage{algpseudocode}
\usepackage{lscape}

\usepackage{subcaption}

\usepackage[numbers]{natbib}

\begin{document}
\setcounter{page}{1}

\vspace*{1.0cm}
\title[RK-GRASP for combinatorial optimization]
{A random-key GRASP for combinatorial optimization}
\author[A.A. Chaves, M.G.C. Resende, R.M.A. Silva]{ A.A. Chaves$^{1,*}$, M.G.C. Resende$^{1,2}$, R.M.A. Silva$^3$}
\maketitle
\vspace*{-0.6cm}

\begin{center}
{\footnotesize {\it

$^1$Departamento de Ciência e Tecnologia, Universidade Federal de São Paulo, São José dos Campos, SP, Brazil\\
$^2$Industrial and Systems Engineering,  University of Washington, Seattle, WA, U.S.A.\\
$^3$Centro de Informática, Universidade Federal de Pernambuco, Recife, PE, Brazil

}}\end{center}

\vskip 4mm {\small\noindent {\bf Abstract.}
This paper proposes a problem-independent GRASP metaheuristic using the random-key optimizer (RKO) paradigm. GRASP (greedy randomized adaptive search procedure) is a metaheuristic for combinatorial optimization that repeatedly applies a semi-greedy construction procedure followed by a local search procedure. The best solution found over all iterations is returned as the solution of the GRASP. Continuous GRASP (C-GRASP) is an extension of GRASP for continuous optimization in the unit hypercube.  A random-key optimizer (RKO) uses a vector of random keys to encode a solution to a combinatorial optimization problem. It uses a decoder to evaluate a solution encoded by the vector of random keys.  A random-key GRASP is a C-GRASP where points in the unit hypercube are evaluated employing a decoder.  We describe random key GRASP consisting of a problem-independent component and a problem-dependent decoder.  As a proof of concept, the random-key GRASP is tested on five NP-hard combinatorial optimization problems: traveling salesman problem, tree of hubs location problem, Steiner triple covering problem, node capacitated graph partitioning problem, and job sequencing and tool switching problem.

\vskip 4mm \noindent {\bf Keywords.}
GRASP; combinatorial optimization; random-key optimizer. 

\footnotetext[1]{Corresponding author. \par E-mail addresses: antonio.chaves@unifesp.br (A. Chaves), mgcr@berkeley.edu (M.G.C. Resende), rmas@cin.ufpe.br (R.M.A. Silva).}


\section{Introduction}
\label{s_introduction}

In this paper, a problem-independent GRASP metaheuristic is proposed using the random-key optimizer (RKO) paradigm 
\cite{Schuetz_RKO_2022}. 
GRASP (greedy randomized adaptive search procedure) is a metaheuristic for combinatorial optimization that repeatedly applies a semi-greedy construction procedure followed by a local search procedure 
\cite{FeoRes89a,Feo_GRASP_1995,ResRib16a}. 
The best solution found over all iterations is returned as the solution of the GRASP. 
Continuous GRASP (C-GRASP) is an extension of GRASP for continuous optimization in the unit hypercube 
\cite{HirMenParRes07a}. 
A random-key optimizer (RKO) makes use of a vector of random keys 
\cite{Bean_RKGA_1994} 
to encode a solution to a combinatorial optimization problem. 
It uses a decoder to evaluate a solution encoded by the vector of random keys. 
A random-key GRASP (RK-GRASP) is a C-GRASP where points in the unit hypercube are evaluated by means of a decoder. 
We describe the random-key GRASP metaheuristic consisting of a problem-independent component and a problem-dependent decoder. 
As a proof of concept, the random-key GRASP is tested on five NP-hard combinatorial optimization problems: traveling salesman problem, tree of hubs location problem, Steiner triple covering problem, node capacitated graph partitioning problem, and job sequencing and tool switching problem.

The paper is organized as follows.
In Section~\ref{RKO}, the concept of Random-Key Optimizer (RKO) and its application in a RK-GRASP are described.
Section~\ref{RK-GRASP} 
describes the implementation of RK-GRASP, including constructive and local search phases.
Section~\ref{s_applications}
presents five combinatorial optimization problems to which RK-GRASP is applied, simply by making use of different decoders.
Experimental results obtained by applying RK-GRASP to the five combinatorial optimization problems are summarized in Section~\ref{s_comp}.
Concluding remarks are made in Section~\ref{s_concl}, where
the key findings and contributions of the paper are summarized.

\section{Random-Key optimizer}
\label{RKO}

Random-key genetic algorithms (RKGA) were first introduced by Bean \cite{Bean_RKGA_1994}.
In an RKGA, solutions are encoded as vectors of random keys, i.e., randomly generated real numbers in the interval $[0,1)$. A population of $p$ random-key vectors is evolved over a number of generations. The initial population consists of randomly generated $n$-vectors. At each iteration, each random-key vector is decoded and evaluated. The \textit{decoder} is a deterministic algorithm, usually a heuristic, that takes as input a vector of random keys and returns a feasible solution of the problem being solved along the cost of the solution. The population is partitioned into an elite set of $p_e$ solutions and a non-elite set of $p-p_e$ solutions. The population is evolved as follows.  First, all elite solutions are copied to the next generation, and $p_m$ mutant solutions, each consisting of an $n$-vector of randomly generated keys, are added to the next generation.
Finally, the remaining $p-p_e-p_m$ elements are made of solutions produced by a crossover of pairs of solutions drawn at random from the population of the current generation.
Mating is done with parameterized uniform crossover \cite{spears1991virtues} where the child inherits the key of one of the parents with probability $\rho$ and of the other with probability $1-\rho$.  We call $\rho$ the \textit{inheritance probability}.
Biased random-key genetic algorithms \cite{Gonçalves_BRKGA_2011} is an extension of RKGA where one of the parents is always selected from the elite set of the population, and that parent is associated with the inheritance probability $\rho > \frac{1}{2}$.

Both RKGA and BRKGA are problem-independent algorithms in the sense that there is a clear separation between the solver and the problem being solved where the connection of the solver with the problem is done by way of a decoder.  For each type of problem to be solved, a new decoder is implemented to serve as the link between the solver and the problem.  These algorithms are examples of \textit{Random-Key Optimizers} (RKO) \cite{Schuetz_RKO_2022}.  An RKO is an optimization heuristic algorithm that solves a discrete optimization problem indirectly in the continuous unit $n$-dimensional hypercube $\mathcal{H}_n$.  For each solution $x \in \mathcal{H}_n$, a decoder $\mathcal{D}$ maps $x$ to the solution $\mathcal{D}(x)$ in the solution space of the discrete optimization problem. With such a separation of solver and decoder, one needs only to implement the solver once and then it can be reused to solve a number of different problems by simply devising a decoder for that problem. Examples of Application Programming Interfaces (APIs) for BRKGA are Toso and Resende \cite{TosRes15a}, Andrade et al. \cite{AndTosGonRes21a}, Oliveira et al. \cite{OliCarOliRes22a}, and Chaves and Lorena \cite{Chaves_BRKGA_QL_2021}.

The first paper to extend the concept of RKO to other algorithmic frameworks was Schuetz et al. \cite{Schuetz_RKO_2022} where in addition to a BRKGA for robot motion planning, the authors propose an RKO using dual annealing, an extension of generalized simulated annealing \cite{TsaSta96a,XiaSunFanGon97a}. Recently, RKO has been implemented in simulated annealing, iterated local search, and variable neighborhood search for the tree hub location problem \cite{ManPolMacJulProGiaSalCha23a}.


\section{Random-Key GRASP}
\label{RK-GRASP}

This section introduces random-key GRASP, or RK-GRASP, a problem-independent GRASP that solves discrete optimization problems through continuous optimization and a decoder.  An advantage of using this GRASP is that the user only needs to implement a decoder since the algorithm-specific components of GRASP are implemented as an API.  When using a standard GRASP the user needs to tailor both the semi-greedy construction and the local search for the problem on hand.
We begin by reviewing the basic GRASP for combinatorial optimization and then consider C-GRASP or continuous GRASP for continuous global optimization over the continuous $n$-dimensional hypercube.

\subsection{GRASP}

The metaheuristic \textit{Greedy Randomized Adaptive Search Procedure} (GRASP) \cite{ResRib16a} was introduced by Feo and Resende \cite{FeoRes89a,Feo_GRASP_1995}.
A GRASP is a multi-start procedure in which at each iteration a semi-greedy solution is constructed and local search is applied to this solution.
In greedy construction, one builds a solution one element at a time.  Among the set of elements whose inclusion in the solution does not lead to infeasibility, the greedy algorithm selects an element that best contributes to the objective function.   In a semi-greedy algorithm, instead of selecting a single element, a set of elements is selected, each having the best or almost the best contribution to the objective function, none of which leads to infeasibility. The set of elements from which to select the single element is called the \textit{restricted candidate list}, or simply RCL. By adjusting the amount of randomness in the selection process, one can mimic a totally random selection or a totally greedy process. Neither the greedy nor semi-greedy construction process is guaranteed to produce a solution that is globally optimal, or even locally optimal with respect to some local neighborhood structure. Given a solution $x$ produced by a greedy or semi-greedy construction procedure, let $\mathcal{N}(x)$ denote the local neighborhood of $x$, i.e. the set of solutions obtained by making a small perturbation (or change) to $x$.  In the case of minimization (maximization), local search stops at local minimum (maximum) $x$ if there is no solution 
$y \in \mathcal{N}(x)$ such the cost $f(y) < f(x)$ (or $f(y) > f(x)$ for the case of maximization).   If such a solution exists, we set $x \gets y$ and the search resumes.  The best locally optimal solution visited over all iterations is returned as the GRASP solution.

\subsection{Continuous GRASP}
\label{CGRASP}
Continuous GRASP, or simply C-GRASP, is an extension of GRASP for solving continuous optimization problems subject to box or simple bounding constraints \cite{HirMenParRes07a,HirParRes10a}, 
$$\min_{x \in \mathbb{R}^n} \{ f(x) \;|\; L_n \leq x \leq U_n \},$$
where $L_n$ and $U_n$ are vectors of lower and upper bounds on $x$, i.e. $L_n(i) \leq x_i \leq U_n(i),$ for $i=1,\ldots,n$ and $f(x)$ is the cost of solution $x$. Cost $f(x)$ can be evaluated in a multitude of ways, e.g. analytically, through simulation, via a mathematical program, or with a decoder. The objective is to find a global optimum. Like GRASP, C-GRASP is a multi-start procedure in which each iteration consists of a construction, or diversification, phase followed by a local search, or intensification, phase. C-GRASP evaluates points on a dynamic grid, with grid size initially set to $h = h_0$. Each construction phase starts at the current solution $x$ (initially a random point $x \in \mathbb{R}^n \;|\; L_n \leq x \leq U_n$). To build an RCL, C-GRASP performs a line search on $f(x)$ in each direction $e_i = (0, 0, \ldots, 0,1, 0, \ldots, 0, 0)$, for $i=1,2,\ldots,n$, where the only nonzero is a 1 in position $i$. For $i=1,2,\ldots,n$, the line search is limited to evaluating $f(x+e_i \cdot h \cdot k)$ for all values of $k \in \{0,1,-1,2,-2, \ldots \}$ such that $L_i \leq x_i+e_i \cdot h \cdot k \leq U_i$. The result of line search $i$ is $z_i$ with cost $g_i = f(z_i)$. Let $g_m = \min \{g_i \;|\; i=1,\ldots,n\}$ and $g_M = \max \{g_i \;|\; i=1,\ldots,n\}$. The best line search solutions, i.e. those with $g_i \leq (1-\alpha) \cdot g_m + \alpha \cdot g_M$ for some $\alpha \in [0,1]$ are placed in an RCL and an index $j$ is selected at random to be fixed in $x$ with $x = z_j$. Direction $e_j$ is
flagged to no longer be explored in this construction iteration. This is repeated until a solution is constructed.  

Once a semi-greedy solution $x$ has been constructed, a local search, or intensification, phase is applied around $x$.
Several implementations of local search have been described. Suppose the current semi-greedy solution
is $\bar{x}$. In the first paper, Hirsch et al. \cite{HirMenParRes07a} suggest examining a given maximum number of points $\texttt{MaxDirToTry}$ of the form $\bar{x} + h \cdot \{ -1,0,1\}^n$.
This local search examines only grid points.
Hirsch, Pardalos, and Resende \cite{HirParRes10a} sample a user-defined number of feasible grid points and project each one of
them onto the surface of the hyper-sphere of radius $h$, centered at $\bar{x}$.  Each point is evaluated as it is projected, and a first-improvement policy is applied.  If an improving point is found, then $\bar{x}$ is set to this improving solution. If no improving solution is found, the grid size is reduced by a specified factor, and the search process continues.
See \cite{HirMenParRes07a,HirParRes10a,SilResParHir13a} for details.

\subsection{Random-Key GRASP} 

We consider optimization problems of the form
$$\min \{ f(\mathcal{D}(x)) \mid x \in \mathcal{H}_n \},$$
where $\mathcal{H}_n$ denotes the unit hypercube in $\mathbb{R}^n$ and $f(\mathcal{D}(x))$ represents the cost value returned by a decoder $\mathcal{D}$ when given $x \in \mathcal{H}_n$ as input.

RK-GRASP, or random-key GRASP, embodies the core principles of both the GRASP and C-GRASP, encompassing constructive and local search phases. This paper introduces a constructive phase inspired by the line search strategy of C-GRASP. Furthermore, we introduce three distinct methods for the local search phase: Grid Search, Nelder-Mead Search, and Random Variable Neighborhood Descent (RVND). Subsequently, we provide the pseudo-codes of these methods and a summary of their concepts.

Algorithm \ref{alg:rk-grasp} presents pseudo-code for RK-GRASP. Initially, a random-key vector $x$ is generated randomly, and the parameter value $h$ is set as $h_s$. The RK-GRASP search process continues while $h$ is greater than or equal to the lower limit ($h_e$). Like GRASP, each iteration consists of constructive and local search phases. Unlike GRASP, the constructive phase depends on the local optimum found in the local search phase of the previous iteration. In the constructive phase, a new solution $x'$ is generated from the current solution $x$ using partially greedy moves. The local search phase intensifies the constructed solution and obtains a local optimal solution $x''$.  If the objective function value of the $x''$ is better than the best solution found so far, $x^*$ is updated. Otherwise, the grid is reduced by halving the value of $h$. \textcolor{black}{Finally, the algorithm evaluates whether the new solution should be accepted. While a straightforward acceptance criterion could be employed, such as accepting all solutions as in C-GRASP, the RK-GRASP uses a simulated annealing acceptance criterion to enhance the solution quality.} The algorithm continues until a stopping criterion is met. 




\begin{algorithm}[htbp]
    \KwData{$n, hs, he$}
    \KwResult{Best solution found during the search process ($x^*$)}
    $f(\mathcal{D}(x^*)) \leftarrow \infty$\;
    \While{not stopping condition}{ 
        $x \leftarrow$ CreateInitialSolution($n$)\; 
        $h \leftarrow hs$\;  
        \While{$h \geq he$}{
            $x' \leftarrow$ ConstructGreedyRandomized($x,h,n$)\; 
            $x'' \leftarrow$ LocalSearch($x',h,n$)\; 
            \eIf{$f(\mathcal{D}(x'')) < f(\mathcal{D}(x^*))$}{ 
                $x^* \leftarrow x''$\;
            }
            { 
                $h \leftarrow h/2$\; 
            }
            \textcolor{black}{\If{accept($x'', x$)}{
                $x \leftarrow x''$\;}
            }
        } 
    } 
    \Return $x^*$ 
    \caption{RK-GRASP}
    \label{alg:rk-grasp}
\end{algorithm}


The constructive phase involves perturbing a solution $x$ with semi-greedy moves that are randomly selected. Algorithm \ref{alg:cgr} presents the pseudocode for the constructive phase. The $\mathit{UnFixed}$ vector is initialized with the indices of all random keys in the solution vector $x$. The constructive phase continues as long as the $\mathit{UnFixed}$ vector is not empty. In each iteration, a LineSearch method is applied to each random key whose index $i$ is in $\mathit{UnFixed}$, returning the best objective function value $g_i$ for that random key obtained through a linear search over the interval $[0,1)$ and its corresponding value $r_i$. Subsequently, the Restricted Candidate List (RCL) is constructed, considering the indices of the random keys that generated solutions with objective function values within the range defined by $\alpha$. The parameter $\alpha$ is randomly set at the beginning of the construction phase within the interval $[0,1]$, where 0 represents the insertion of only the index of the best random key, and 1 represents the insertion of all indices that are $\mathit{UnFixed}$. From the RCL, an index $j$ is randomly selected, the random key value of this index is updated with the value found by LineSearch, and this index is removed from $\mathit{UnFixed}$. To speed up the construction process \cite{HirParRes10a}, we use the $\mathit{Reuse}$ flag to bypass the application of the LineSearch in subsequent iterations where the random key has not been modified (i.e., when $x_j = r_j$).

\begin{algorithm}[htbp]
    \KwData{$x,h,n$}
    \KwResult{A constructive semi-greedy solution.}
    $\mathit{UnFixed} \leftarrow \{1,2,...,n\}$\; 
    $\alpha \leftarrow$ {\fontfamily{pcr}\selectfont UnifRand(0,1)}\;
    $\mathit{Reuse} \leftarrow \mathit{false}$\;
    \While{$\mathit{UnFixed} \neq \emptyset$}{ 
        $min \leftarrow +\infty$; $max \leftarrow - \infty$\; 
        \For{$i = 1,..., n$}{ 
            \If{$i \in \mathit{UnFixed}$}{ 
                \If{$\mathit{Reuse} = \mathit{false}$}{
                    $[r_i, g_i] \leftarrow$ LineSearch($x,h,i$)\; 
                }
                \If{$min > g_i$}{$min \leftarrow g_i$\;}
                \If{$max < g_i$}{$max \leftarrow g_i$\;}
            }
        }
        $RCL \leftarrow \emptyset$\;
        \For{$i = 1,..., n$}{ 
            \If{$i \in \mathit{UnFixed} \wedge g_i \leq min + \alpha \cdot (max - min)$}{ 
                $RCL \leftarrow RCL \cup \{i\}$\; 
            }
        }
        $j \leftarrow$ RandomlySelectElement($RCL$)\;
        \eIf{$x_j = r_j$}{
            $Reuse \leftarrow true$\;
        }
        {
            $x_j\leftarrow r_j$\; 
            $f(\mathcal{D}(x)) \leftarrow g_j$\;
            $\mathit{Reuse} \leftarrow \mathit{false}$\;
        }
        $\mathit{UnFixed} \leftarrow \mathit{UnFixed} \backslash \{j\}$\; 
    } 
    \Return $x$ 
    \caption{ConstructGreedyRandomized}
    \label{alg:cgr}
\end{algorithm}

The LineSearch begins by generating a set of integers $K$ within the range $[-1/h, 1/h]$. The next step involves generating a set of feasible random keys $RK$ derived from the initial random key $x_i$ and the values $K_j$ and $h$ ($0 \leq x_i + K_j \cdot h < 1$). The random key $i$ is then updated with each value contained in $RK$, and the solution $x$ is decoded to evaluate its objective function value. In this paper, we propose adapting the greedy algorithm so that instead of selecting the best options, it only considers a subset of $q < |RK|$ possible random-key values (chosen randomly) during each iteration. LineSearch returns, from this subset, the random-key value that generates the solution with the best objective function value ($\mathit{best}_r$ and $\mathit{best}_f$).

\begin{algorithm}[htbp]
    \KwData{$x,h,i$}
    \KwResult{The best value for random key $i$.}
    $K \leftarrow \{0,1,-1,2,-2,...,1/h,-1/h\}$\; 
    $RK \leftarrow \emptyset$\;
    \For{$j = 1,..., |K|$}{ 
        \If{$0 \leq (x_i + K_j \cdot h) < 1$}{
            $RK \leftarrow RK \cup (x_i + K_j \cdot h)$\;
        }
    }
    $q \leftarrow \lceil log_2 (1/h) \rceil$ + 1\; 
    $\mathit{best}_f \leftarrow \infty$; $\mathit{best}_r \leftarrow 0$\;
    \For{$t = 1,..., q$}{ 
        $j \leftarrow$ {\fontfamily{pcr}\selectfont UnifRand(1,$|RK|$)}\;
        $x_i \leftarrow RK_{j}$\; 
        \If{$f(\mathcal{D}(x)) < \mathit{best}_f$}{
            $\mathit{best}_f \leftarrow f(\mathcal{D}(x))$\;
            $\mathit{best}_r \leftarrow x_i$\;
        }
        $RK \leftarrow RK \backslash \{j\}$
    }
    \Return $\mathit{best}_r, \mathit{best}_f$ 
    \caption{LineSearch}
    \label{lineSearch}
\end{algorithm}

The Local Search phase intensifies the search in the neighborhood of the constructed solution, seeking improved solutions. \textcolor{black}{RK-GRASP can incorporate different local search strategies. In this paper, we developed three local search heuristics: Grid Search, Nelder-Mead Search, and Random Variable Neighborhood Descent (RVND). These local search heuristics are problem-independent and intensify the search within the random-key solution space, utilizing the decoder to evaluate neighboring solutions.}



Algorithm \ref{alg:localGrid} presents pseudo-code for the Grid Search heuristic \cite{hirsch2011correspondence}. This heuristic generates a neighborhood and determines at which points in the neighborhood, if any, the objective function improves. If an improvement point is found, it is made the current point, and the local search continues with the new solution. We use the Algorithm \ref{hNeighborhood-Code} to create a solution in the $h$ neighborhood of the current solution $\bar{x}$. Consider the set of points $v_i$ that are integer steps (of size $h$) away from $\bar{x}$. The $h$-Neighborhood algorithm defines the projection of these points onto the hyper-sphere of radius $h$, centered at $\bar{x}$ . We use the parameter $\theta$ to control the intensity of this projection, maintaining a random key fixed with a probability of $1 - \theta$. An $h$-neighbor of the solution $\bar{x}$ is defined by the current random keys, $h$, $v$, and the norm of $v-\bar{x}$. The algorithm randomly selects solutions in the $h$-neighborhood, one at a time. 

\begin{algorithm}[htbp]
    \KwData{$x,h,n,\theta$}
    \KwResult{The best neighbor found in the grid of solution $x$.}
    $\bar{x} \leftarrow x$\;
    $NumGridPoints \leftarrow n \cdot \lfloor 1.0/h \rfloor$\;
    $\mathit{NumPointsExamined} \leftarrow 0$\;
    \While{$\mathit{NumPointsExamined} \leq \mathit{NumGridPoints}$}{
        $\mathit{NumPointsExamined} \leftarrow \mathit{NumPointsExamined} + 1$\;
        $y \leftarrow \mathit{h}\mathrm{Neighborhood}(\bar{x},h,n,\theta)$\;
            \If{$f(\mathcal{D}(y)) < f(\mathcal{D}(\bar{x}))$}{
                $\bar{x}\leftarrow y$\;
                $\mathit{NumPointsExamined} \leftarrow 0$\;
            }
    }
    \Return $\bar{x}$ 
    \caption{GridSearch}
    \label{alg:localGrid}
\end{algorithm}


\begin{algorithm}[htbp]
    \KwData{$\bar{x},h,n,\theta$}
    \KwResult{A neighbor in the grid of solution $\bar{x}$.}
    \For{$i = 1,..., n$}{ 
        \eIf{$\textrm{{\fontfamily{pcr}\selectfont UnifRand}}(0, 1) \leq 0.5$}{ 
           $v_i \leftarrow  \left \lceil \textrm{{\fontfamily{pcr}\selectfont UnifRand}}(1, (1.0 - \bar{x}_i) / h) \right \rceil$\; 
        }
        {
            $v_i \leftarrow  -1 \cdot \left \lceil \textrm{{\fontfamily{pcr}\selectfont UnifRand}}(1, \bar{x}_i / h) \right \rceil$\; 
        }
    }
    \textcolor{black}{\For{$i = 1,..., n$}{
        \If{$\textrm{{\fontfamily{pcr}\selectfont UnifRand}}(0, 1) \leq \theta$}{
            $x_i \leftarrow \bar{x}_i + h \cdot (v_i / \left \| v \right \|)$\; 
        }}
    }
    \Return $x$ 
    \caption{$h$Neighborhood}
    \label{hNeighborhood-Code}
\end{algorithm}

            


The Nelder–Mead method, introduced by Nelder and Mead \cite{nelder1965simplex}, is a numerical technique for searching the minimum of an objective function within a multidimensional space. This direct search approach, relying on function comparisons, is commonly utilized in derivative-free nonlinear optimization. This method begins with at least three solutions and can perform five moves: reflection, expansion, inside contraction, outside contraction, and shrinking. In this study, we always apply the Nelder-Mead Search with three solutions: $x_1$, $x_2$, and $x_3$, where one is the current solution derived from the constructive phase, while the others are randomly chosen from a pool of elite solutions found during the search process. These solutions are ordered by objective function value ($x_1$ is the best and $x_3$ is the worst). Figure \ref{fig:NMexample} shows an illustrative example of a simplex polyhedron and the five moves.

\begin{figure}[htbp]
    \centering
    \includegraphics[width=0.85\textwidth]{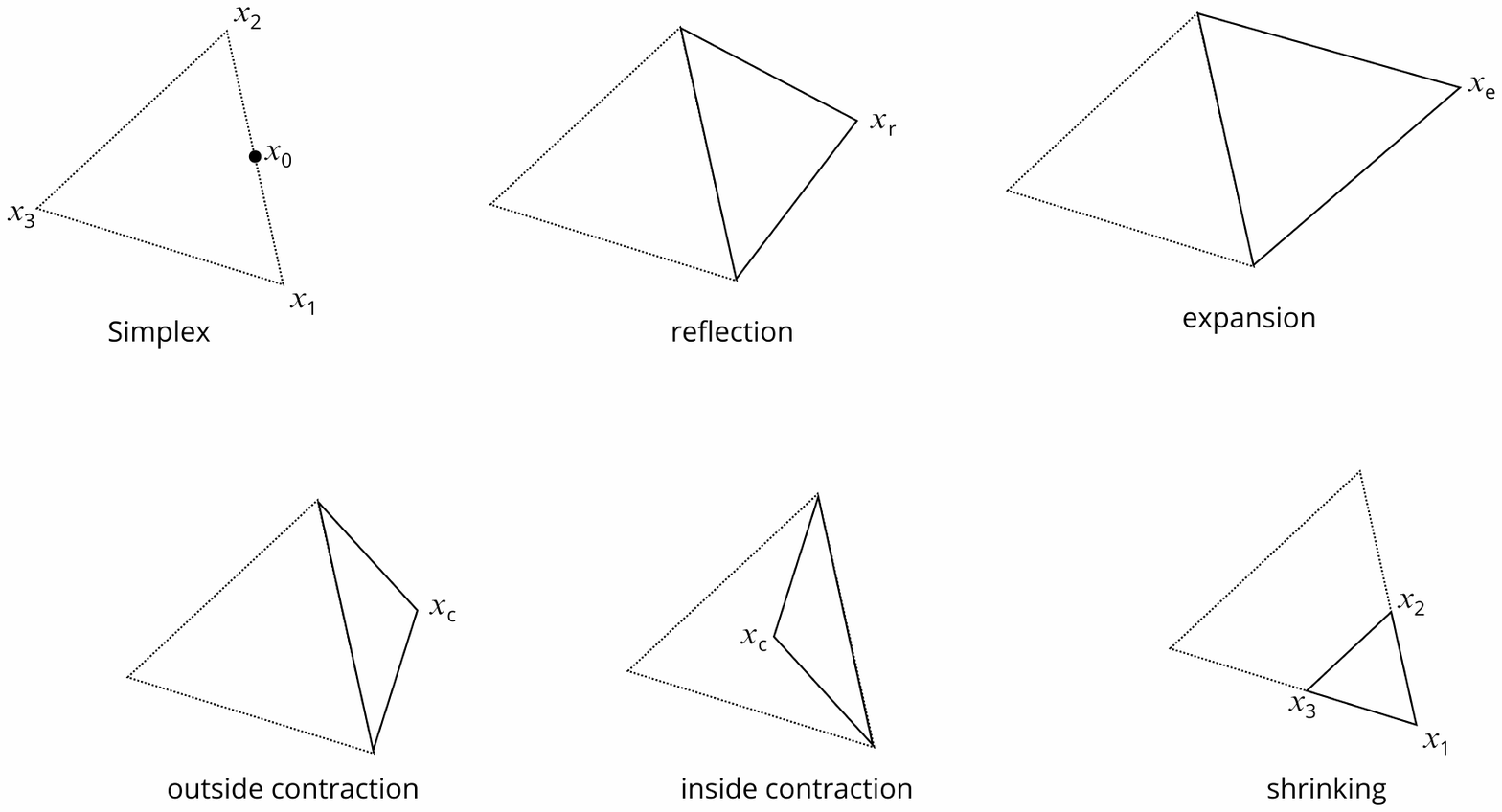}
    \caption{Illustrative example of the simplex polyhedron and the five moves of the Nelder-Mead Search. Source: based on the figure in \cite{kolda2003optimization}}
    \label{fig:NMexample}
\end{figure}

Algorithm \ref{alg:neldermead} presents pseudo-code for the Nelder-Mead Search adapted for discrete optimization problems. We use an adapted Uniform Crossover (UX) \cite{davis1989handbook} to generate the moves. In the UX between solutions $a$ and $b$, for each random key of the vector, we decide at random from which solution the random key is copied to the corresponding random key of the new solution. We incorporate a parameter $\mathit{factor}$ to determine whether, when using the random key of solution $b$, to use the value of the random key itself ($\mathit{facto}r=1$) or an inverted value of the random key ($\mathit{factor}=-1$).

The algorithm begins with an initial simplex of three solutions ($x_1,x_2,x_3$). The simplex is sorted based on the objective function values, and the simplex's centroid ($x_0$) is computed between $x_1$ and $x_2$ $(x_0=$\texttt{UX}$(x_1,x_2,1))$. The main loop iterates until a termination condition is met. The algorithm performs a series of moves on the simplex during each iteration to explore the search space. A reflection solution $(x_r = $\texttt{UX}$(x_0, x_3, -1))$ is computed. If the objective function value at $x_r$ is better than the current best solution ($x_1$), the algorithm computes an expansion solution $(x_e = $\texttt{UX}$(x_r, x_0, -1))$. If the objective function value at $x_e$ is better than at $x_r$, $x_3$ is replaced by $x_e$; otherwise, $x_3$ is replaced by $x_r$. If neither the reflection nor the expansion improves the solution, the algorithm contracts towards the solution $x_r$ or $x_3$. For an outside contraction (when $x_r$ is better than $x_3$), the contraction solution is $x_c = $\texttt{UX}$(x_r, x_0, 1)$. For an inside contraction (when $x_r$ is not better than $x_3$), the contraction solution is $x_c = $\texttt{UX}$(x_0, x_3, 1)$. If the contraction step does not improve, the entire simplex is shrunk towards the best solution $x_1$ $(x_i = $\texttt{UX}$(x_1, x_i, 1), i=2,3)$. The algorithm terminates when a specified termination condition is met. In this paper, we use the same criterion of the Grid Search, the maximum number of iterations equal $n \cdot \lfloor 1.0/h \rfloor$.

\begin{algorithm}[htbp]
    \KwData{$x_1$, $x_2$, $x_3$, $n$, $h$}
    \KwResult{The best solution found in \textit{simplex} $X$.}
    Initialize simplex: $X \leftarrow \{x_1, x_2, x_3\}$\;
    Sort simplex $X$ by objective function value\;
    Compute the simplex centroid $x_0 \leftarrow$  \texttt{UX}$(x_1,x_2,1)$\;
    $\mathit{iter} \leftarrow 0$\; 
    $numIter \leftarrow n \cdot \lfloor 1.0/h \rfloor$\;
    \While{$\mathit{iter} < \textit{numIter}$}{
        $shrink \leftarrow 0$\; 
        $\mathit{iter} \leftarrow \mathit{iter} + 1$\; 
        Compute reflection solution 
        $x_r \leftarrow$  \texttt{UX}$(x_0,x_3,-1)$\;
        \If{$f(\mathcal{D}(x_r)) < f(\mathcal{D}(x_1))$}{
            Compute expansion solution  $x_e \leftarrow$  \texttt{UX}$(x_r,x_0,-1)$\;
            \If{$f(\mathcal{D}(x_e)) < f(\mathcal{D}(x_r))$}{
                $x_3 \leftarrow x_e$\;
            }
            \Else{
                $x_3 \leftarrow x_r$\;
            }
        }
        \Else{
            \If{$f(\mathcal{D}(x_r)) < f(\mathcal{D}(x_2))$}{
                $x_3 \leftarrow x_r$\;
            }
            \Else{
                \If{$f(\mathcal{D}(x_r)) < f(\mathcal{D}(x_3))$}{
                    Compute contraction solution  $x_c \leftarrow$  \texttt{UX}$(x_r,x_0,1)$\;
                    \If{$f(\mathcal{D}(x_c)) < f(\mathcal{D}(x_r))$}{
                        $x_3 \leftarrow x_c$;\
                    }
                    \Else{
                        $shrink \leftarrow 1$\;
                    }
                }
                \Else{
                    Compute contraction solution  $x_c \leftarrow$  \texttt{UX}$(x_0,x_3,1)$\;
                    \If{$f(\mathcal{D}(x_c)) < f(\mathcal{D}(x_3))$}{
                        $x_3 \leftarrow x_c$;\
                    }
                    \Else{
                        $shrink \leftarrow 1$\;
                    }
                }
            }
        }
        \If{$shrink = 1$}{
            Replace all solutions except the best $x_1$ with  $x_i \leftarrow$ \texttt{UX}$(x_1,x_i,1), i=2,3$\;
        }
        Sort simplex $X$ by objective function value\;
        Compute the simplex centroid  $x_0 \leftarrow$  \texttt{UX}$(x_1,x_2,1)$\;
    }
    \Return $x_1$ 
\caption{NelderMeadSearch}\label{alg:neldermead}
\end{algorithm}

Since a local optimal for a given local search heuristic may not be a local optimal for another local search heuristic, using different heuristics usually leads to a better improvement phase. However, the order in which local search heuristics are explored must be determined. For this reason, we use a variation of the Variable Neighborhood Descent (VND) \cite{Mladenovic_VNS_1997} search that does not have a predefined order for the heuristics to be applied. This search is called Random Variable Neighborhood Descent (RVND) \cite{Penna_RVND}. RVND randomly selects the neighborhood heuristic order to be applied in each iteration. Algorithm \ref{alg:RVND} shows pseudo-code for the RVND.

The RVND algorithm starts by initializing a list of neighborhoods ($NL$) representing different local search heuristics to explore the solution space. It iteratively selects a neighborhood $\mathcal{N}^i$ randomly and searches for the best neighboring solution ($x'$) from that neighborhood. If the new solution is better than the current solution ($x$), it updates $x$ with $x'$ and restarts the exploration of neighborhoods. Otherwise, the current neighborhood will be removed from the list. This process continues until the list of neighborhoods is empty. Finally, the algorithm returns the best solution found. RVND efficiently explores diverse solution spaces but can also be applied to random-key spaces. Users can implement classic heuristics for the problem on hand and encode the local optimal solution into the random-key vector after the search process or implement random-key neighborhoods independent of the problem on hand using the decoder to converge towards better solutions.

\begin{algorithm}[htbp]
    \KwData{$x$}
    \KwResult{The best solution in the neighborhoods.}
    Initialize the Neighborhood List ($NL$)\;
    \While{$NL \neq 0$}{
        Choose a neighborhood $\mathcal{N}^i \in NL$ at random\;
        Find the best neighbor $x'$ of $x \in \mathcal{N}^i$\;
        \If{$f(\mathcal{D}(x')) < f(\mathcal{D}(x))$}{
            $x \leftarrow x'$\;
            Restart $NL$\;
        }
        \Else{
            Remove $\mathcal{N}^i$ from the $NL$ \;
        }
    }
    \Return $x$
\caption{RandomVND} \label{alg:RVND}
\end{algorithm}

\textcolor{black}{In this paper, we developed three additional problem-independent local search heuristics that operate on the random-key vector. These heuristics encompass distinct neighborhood structures: SwapRK, InvertRK, and FareyRK. 
The SwapRK structure focuses on interchanging two positions within the random-key vector. Conversely, the InvertRK structure perturbs the random-key values, inverting the current value ($1 - x_i$). 
Lastly, the FareyRK structure randomly changes the current value of each random key with random values generated between each two consecutive terms of the Farey sequence \cite{niven1991introduction}. 
The Farey sequence of order $\eta$ is the sequence of completely reduced fractions between 0 and 1, with denominators less than or equal to $\eta$, arranged in increasing order. We use the Farey sequence of order 7,
$$F_7 = \left\{ 
\frac{0}{1}, 
\frac{1}{7}, 
\frac{1}{6}, 
\frac{1}{5}, 
\frac{1}{4}, 
\frac{2}{7}, 
\frac{1}{3}, 
\frac{2}{5}, 
\frac{3}{7}, 
\frac{1}{2}, 
\frac{4}{7}, 
\frac{3}{5}, 
\frac{2}{3}, 
\frac{5}{7}, 
\frac{3}{4}, 
\frac{4}{5}, 
\frac{5}{6}, 
\frac{6}{7}, 
\frac{1}{1} 
\right\}$$
for our purposes.} The random keys are explored in random order in each iteration of these heuristics. \textcolor{black}{The Grid Search, the Nelder-Mead Search, the SwapRK, the InvertRK, and FareyRK heuristics are the neighborhoods of the RVND algorithm, which systematically explores all neighbors of the current solution using the first-improvement strategy, i.e., any improving move is immediately applied.}


\section{Applications}
\label{s_applications}

This section presents the applications of RK-GRASP to five combinatorial optimization problems:

\begin{enumerate}
    \item Traveling salesman problem
    \item Hub location problem in graphs
    \item Steiner triple set covering problem
    \item Node capacitated graph partitioning problem
    \item Job sequencing and tool switching problem
\end{enumerate}

We present a concise overview of each problem, including descriptions of the solution encoding and a corresponding decoder. We aim to illustrate the wide applicability of RK-GRASP as a problem-independent approach.



Initially, we directed our efforts toward applying the RK-GRASP to the Traveling Salesman Problem (TSP). Our rationale behind this choice was establishing an application in a widely recognized and classic problem domain. By doing so, we aimed to facilitate the adoption of the RK-GRASP among new users, leveraging the familiarity and accessibility of the TSP as a benchmark. The framework of the RK-GRASP applied to solve the TSP is available in \url{www.github.com/antoniochaves19/RK-GRASP}. Other problems reuse this framework with the specific problem-dependent component (read data and decoder).

\newpage
\subsection{Traveling Salesman Problem.}

\subsubsection{Problem definition.}

The Traveling Salesman Problem (TSP) \cite{dantzig1954solution} is a classic optimization problem in which the objective is to find the shortest possible route that visits each city exactly once and returns to the original city. Mathematically, given a set of cities ($N$) and the distances between them, the goal is to determine the optimal permutation of cities that minimizes the total distance traveled by the salesman. The TSP is a well-studied combinatorial optimization problem with logistics, transportation, and network design applications. We recommend reading \cite{reinelt2003traveling,gutin2006traveling} to review the literature on the TSP.

\subsubsection{Solution encoding.} Each solution is encoded as a vector $x$ of random keys of length $n = |N|$, where the \textit{i}-th random key corresponds to the \textit{i}-th city of $N$.

\subsubsection{Solution decoder.}
We use the best decoder developed for the TSP \cite{Chaves_BRKGA_QL_2021} to map the random key and TSP solutions. The decoder starts by ordering the random-key vector in ascending order. Then, using this sequence, the cheapest insertion constructive heuristic is performed. A sub-route is created with the first three cities, and from the fourth city in the sequence, the cheapest position is calculated to insert the current city. The cost of inserting a city $k$ after city $j$ is calculated by $c_{k, j} = d_{j, k} + d_{k, j + 1} - d_{j, j + 1}$, where $d_{i,j}$ is the distance or costs between the cities $i$ and $j$. In the end, the cost of the TSP route is computed. Figure \ref{fig:decoderTSP} shows an example of this decoder with five cities.

\begin{figure}[htbp]
    \centering
    \includegraphics[width=0.6\linewidth]{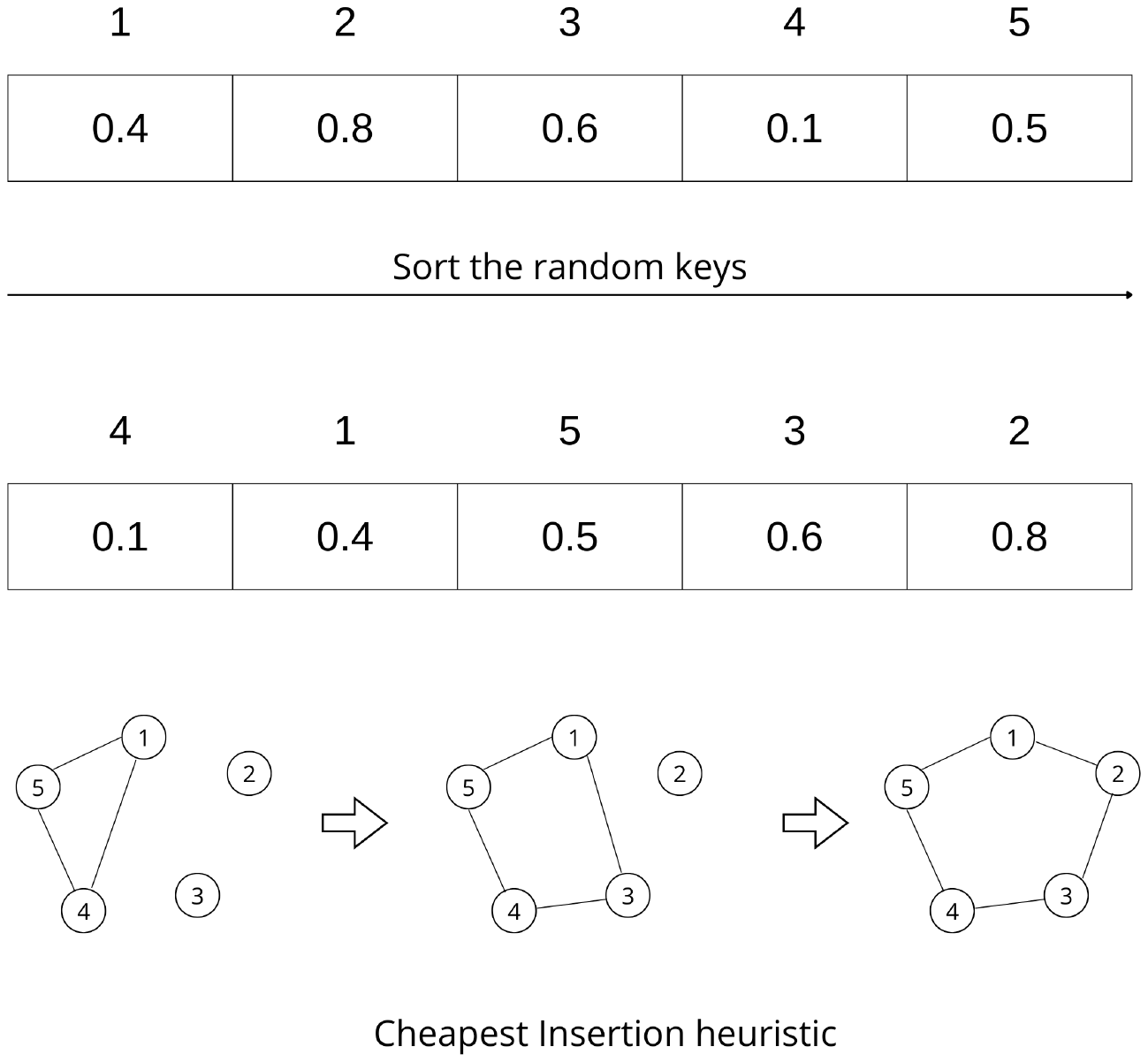}
    \caption{Example of the TSP decoder with five cities.}
    \label{fig:decoderTSP}
\end{figure}







\subsection{Tree Hub Location Problem.}


\subsubsection{Problem definition.} The Tree Hub Location Problem (THLP) \cite{contreras2010} consists of defining $p$ \textit{hubs} in a network that are connected by a tree with non-directional arcs. From this tree, each non-\textit{hub} node must be connected to a \textit{hub} so that all the flow between the nodes must use the \textit{hubs} to travel. Each arc has a transportation cost per unit, and we have a demand between each pair of nodes in the network. The objective function involves two components: firstly, computing the transportation costs for fulfilling all demands from non-hub nodes to a designated hub, and secondly, factoring in the costs associated with the flow within the hub tree. A discount factor is applied to achieve this, allowing for a consideration of the overall cost dynamics involved in the transportation process.

The literature on THLP encompasses both exact approaches, such as Mixed Integer Linear Programming (MILP) models \cite{contreras2009tight, contreras2010} and Benders decomposition \cite{de2013improved}, as well as heuristic methods like BRKGA \cite{pessoa2017biased}. Additionally, some variants of THLP have been proposed \cite{de2015exact,kayicsouglu2021multiple}, expanding the scope of the problem and its solution space. Notably, BRKGA has emerged as the state-of-the-art method for THLP, presenting the best results reported in the literature.

\subsubsection{Solution encoding.}

In this study, we adopt the encoder proposed by Pessoa \textit{et al}. \cite{pessoa2017biased} for our analysis. Consequently, the THLP solution is denoted by a random-key vector composed of three distinct segments. As depicted in Figure \ref{fig:decoderTHLP}, the first segment of the vector, with a dimensionality of $|N|$, signifies the network nodes. The second segment, sized at $|N| - p$, assigns non-hub nodes to their corresponding hubs. Finally, the third segment, sized at $p(p-1)/2$, represents all possible pairs of edges connecting the hubs.

\subsubsection{Solution decoder.}

Our decoder is also based on the decoder presented in \cite{pessoa2017biased}. It begins by sorting the vector’s first segment to delineate hubs and non-hubs: the initial $p$ positions denote hubs, while the subsequent $|N|-p$ positions signify non-hubs. Subsequently, the second segment undertakes the task of assigning non-hub nodes to their respective hubs. Achieving this involves partitioning the interval [0, 1] into $p$ equal intervals, each associated with a specific hub per the preceding ordering. Lastly, the decoder arranges the random keys corresponding to inter-hub arcs in ascending order. The resulting tree structure is then constructed using the Kruskal algorithm applied to the sorted arcs. Figure \ref{fig:decoderTHLP} shows an example of the THLP decoder with $|N|=10$ and $p=3$.

\begin{figure}[htbp]
    \centering
    \includegraphics[width=0.8\linewidth]{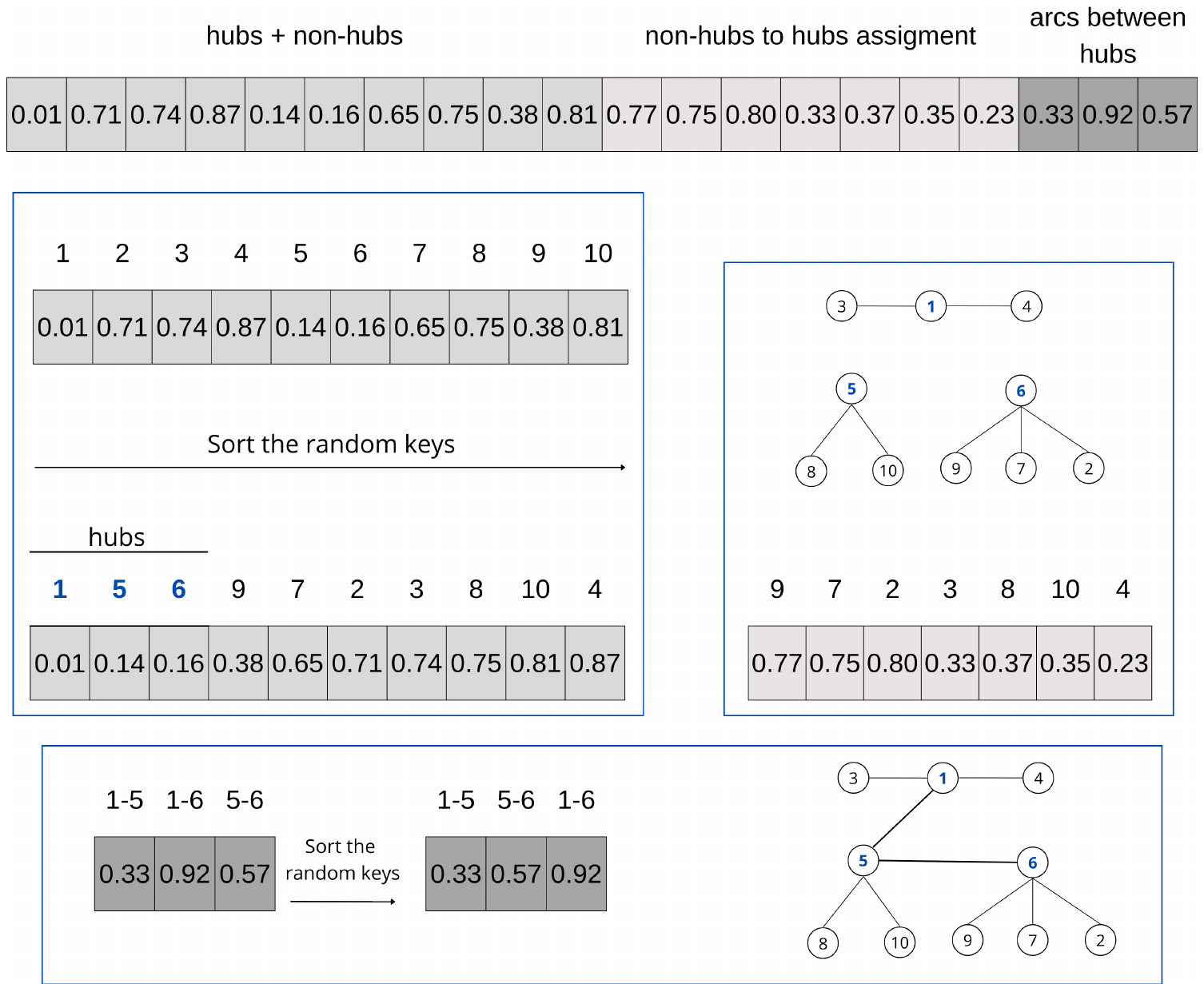}
    \caption{Example of the THLP decoder with ten points and three hubs.}
    \label{fig:decoderTHLP}
\end{figure}




\subsection{Job Sequencing and Tool Switching Problem.} 


\subsubsection{Problem definition.}

The Job Sequencing and Tool Switching problem (SSP) \cite{tang1988} considers scheduling a set of jobs ($T$) to be executed on a single machine. Each job $t$ requires a particular set of tools ($F_t$) for processing, with the job being executable only when this tool subset is available on the machine. However, tool switches become critical when the machine is limited to holding a maximum of C tools simultaneously, and C is less than the total number of tools required for all jobs.

In the SSP, the objective is to sequence jobs to minimize the total number of tool switches. A tool switch occurs when a tool is replaced with another in the machine's magazine. To put it simply, consider you have a toolbox with a limited number of compartments. Each compartment can hold a different tool. Now, if you need to use a tool that is not currently in the toolbox, you must remove a tool from one of the compartments to make space for the new tool. 

The SSP is a \textit{NP}-hard problem (\cite{crama1996minimizing, tang1988}), in which we have two interconnected decision processes: determine the optimal job processing sequence and identify the necessary tools for each job's processing to minimize tool switches, given a fixed job processing sequence. For the last decision, Tang \textit{et al}. \cite{tang1988} introduce an optimal policy called KTNS (Keep Tool Needed Sooner). According to this policy, when inserting a new tool and removing another from the magazine, priority is given to retaining the tools required for upcoming job tasks.

The literature review on the SSP highlights the research focus on heuristic and metaheuristic strategies due to limited exact solution approaches. Mathematical models are proposed by \cite{tang1988, bard1988heuristic, crama1996minimizing, privault2000k, laporte2004exact, ghiani2010solving, da2021new} with several advances in terms of efficiency for solving SSP instances. The literature also presents novel heuristics, such as Tabu Search \cite{al2003tabu} and Beam Search \cite{zhou2005beam}. Furthermore, advancements like memetic algorithms by \cite{amaya2020deep} and graph-based representations \cite{paiva2017improved} offered promising avenues for SSP solutions. \cite{chaves2016hybrid} proposed hybrid methods combining clustering search and BRKGA, while \cite{mecler2021simple} presented a hybrid genetic search, marking ongoing progress in SSP solution methodologies. We recommend reading \cite{da2021new} and \cite{mecler2021simple} for more in-depth literature reviews about SSP.


\subsubsection{Solution encoding.}

We encode a solution of the SSP as a vector of $n$ random keys, where $n = |T|$ is the number of jobs and the $i$-th random key represents the $i$-th job.

\subsubsection{Solution decoder.}

 Decoding the $n$ random keys of each solution into a sequence of jobs is accomplished by sorting the jobs in ascending order of their corresponding random key values. Figure \ref{fig:decoderMTSP} shows an example of the decoding process for the SSP. In this example, there are five jobs. The sorted random keys correspond to the jobs sequence (1, 4, 3, 2, 5). The objective function value of a solution with the job sequence is obtained with the KTNS algorithm \cite{tang1988}. It returns the number of tool switches needed to process the jobs in this sequence. 

\begin{figure}[htbp]
    \centering
    \includegraphics[width=0.75\linewidth]{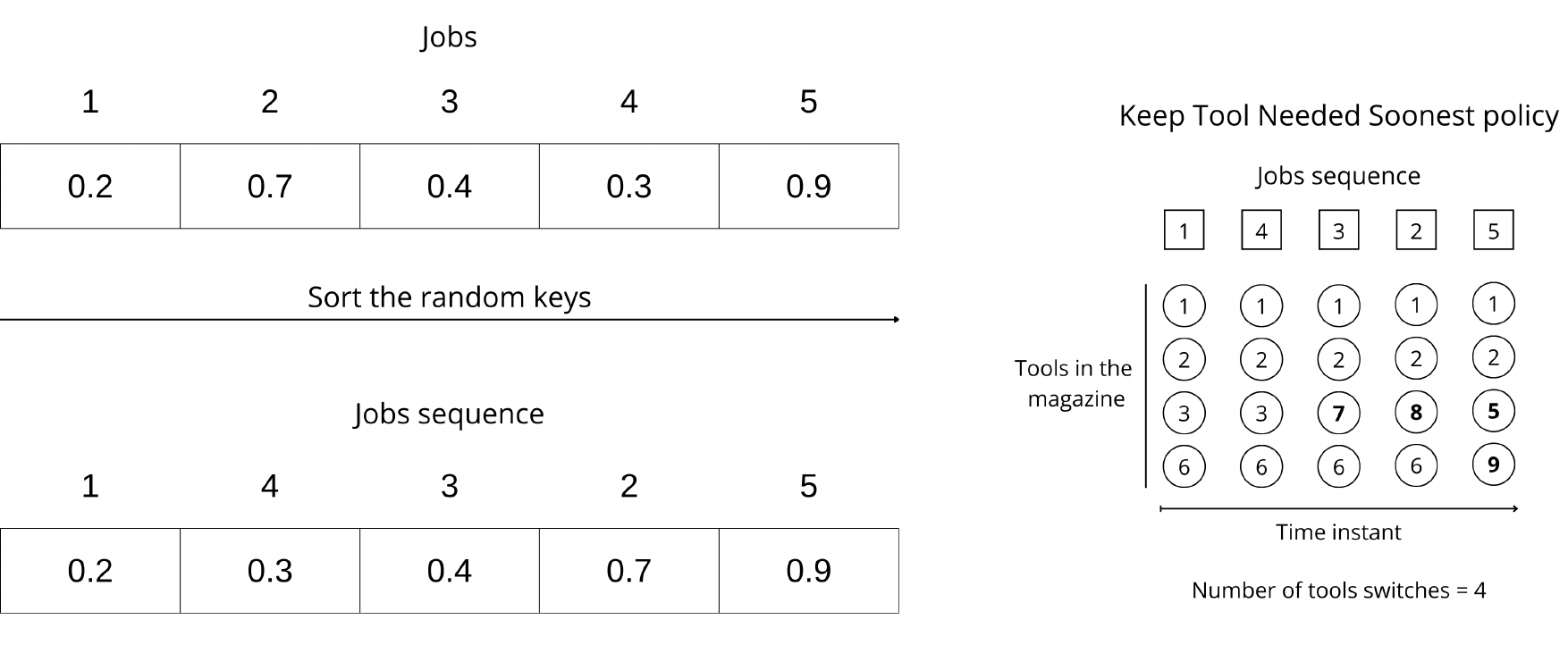}
    \caption{Example of the SSP decoder with five jobs and a magazine with four slots.}
    \label{fig:decoderMTSP}
\end{figure}



\newpage
\subsection{Steiner Triple Covering Problem.} 


\subsubsection{Problem definition.}




Fulkerson et al. \cite{fulkerson1974} introduced the Steiner Triple Covering Problem as a computationally challenging set covering problem category.
A Steiner triple system $B$ on a set $X$ with $N$ elements is a collection of triples (subsets of $X$ with three elements), such that any pair of elements $x$ and $y$ from $X$ is contained in exactly one triple of $B$. For example, the Fano plane (Figure \ref{fig:fano}) exemplifies a Steiner triple system on a set $X$ with seven elements, where the triples represent the seven lines, each containing three points, ensuring that every pair of points is part of a unique line.

A Steiner triple system can be represented by a binary incidence matrix $A$, which has a column for every element in $X$ and a row for each triple in $B$. In matrix $A$, the entry $A(i,j)$ is equal to 1 if and only if the $j$-th element is part of the $i$-th triple. Consequently, every row $i$ in matrix $A$ will have precisely three entries where $A(i,j)$ equals 1. Furthermore, for every pair of columns $j$ and $k$ there is exactly one row $i$ for which $A_{i,j} = A_{i,k} = 1$.

\begin{figure}[htpb]
    \centering
    \begin{subfigure}{0.45\textwidth}
        \centering
        \includegraphics[width=0.5\textwidth]{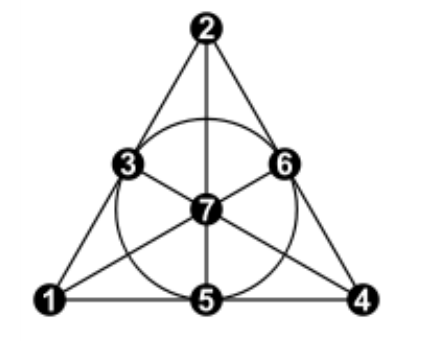}
        \caption{Fano Plane with $X = \{1,\ldots,7\}$, and $B = \{\{1, 2, 3\},\{2,4,6\},\{1,4,5\},\{1,6,7\},$ $\{3,4,7\},\{2,5,7\}\}$.}
        \label{fig:fano1}
    \end{subfigure}
    \hfill
    \begin{subfigure}{0.45\textwidth}
        \centering
        $\begin{bmatrix}
            1 & 1 & 1 & 0 & 0 & 0 & 0\\
            0 & 1 & 0 & 1 & 0 & 1 & 0\\
            1 & 0 & 0 & 1 & 1 & 0 & 0\\
            1 & 0 & 0 & 0 & 0 & 1 & 1\\
            0 & 0 & 1 & 1 & 0 & 0 & 1\\
            0 & 1 & 0 & 0 & 1 & 0 & 1 \\
        \end{bmatrix}$
        \caption{Incidence matrix of Fano Plane.}
        \label{fig:matrixfano1}
    \end{subfigure}
    \caption{Example of STCP input data.}
    \label{fig:fano}
\end{figure}



Several heuristic methods for addressing the Steiner triple covering problem have been suggested. Feo and Resende \cite{FEO198967} introduced a GRASP heuristic, through which they identified a cover of size 61 for $A_{81}$, a result later confirmed as optimal by Mannino and Sassano \cite{MANNINO19951}. Karmarkar et al. \cite{KMR1991} developed an interior point method that achieved a cover of size 105 for $A_{135}$. In the same study, they also applied a GRASP heuristic, finding an improved cover of size 104, a result matched by Mannino and Sassano \cite{MANNINO19951}. In 1998, Odijk and van Maaren \cite{OdijkM98} produced a cover of size 103. This result was recently validated as optimal by Ostrowski et al. \cite{OLRS2009,OLRS2010} and Östergård and Vaskelainen \cite{OV2011}. Moreover, Östergård and Vaskelainen \cite{OV2011} identified all possible solutions for $A_{135}$, demonstrating their isomorphism.

\subsubsection{Solution encoding.}

We encoded a solution of the STCP as a random-key vector of length $n = |X|$, where the $i$-th random key corresponds to the $i$-th element of $X$ (a column of $B$).

\subsubsection{Solution decoder.}

The decoder process for the STCP is designed to create an initial solution and improve it by eliminating redundancy. The random-key vector is sorted and a loop iterates through the column sequence, covering the necessary rows and tracking the number of used columns and coverage redundancy. If a column is used to cover at least one row, it is marked as used. After ensuring all rows are covered, the function attempts to reduce redundancy by checking each used column to see if it is necessary for maintaining coverage. Redundant columns are removed, and the coverage vector is updated accordingly. Finally, we calculates the objective function value, representing the number of columns used. Figure \ref{fig:decoderSTCP} shows an example of the STCP decoder considering the Fano Plane.

\begin{figure}[hbpt]
    \centering
    \includegraphics[width=0.75\linewidth]{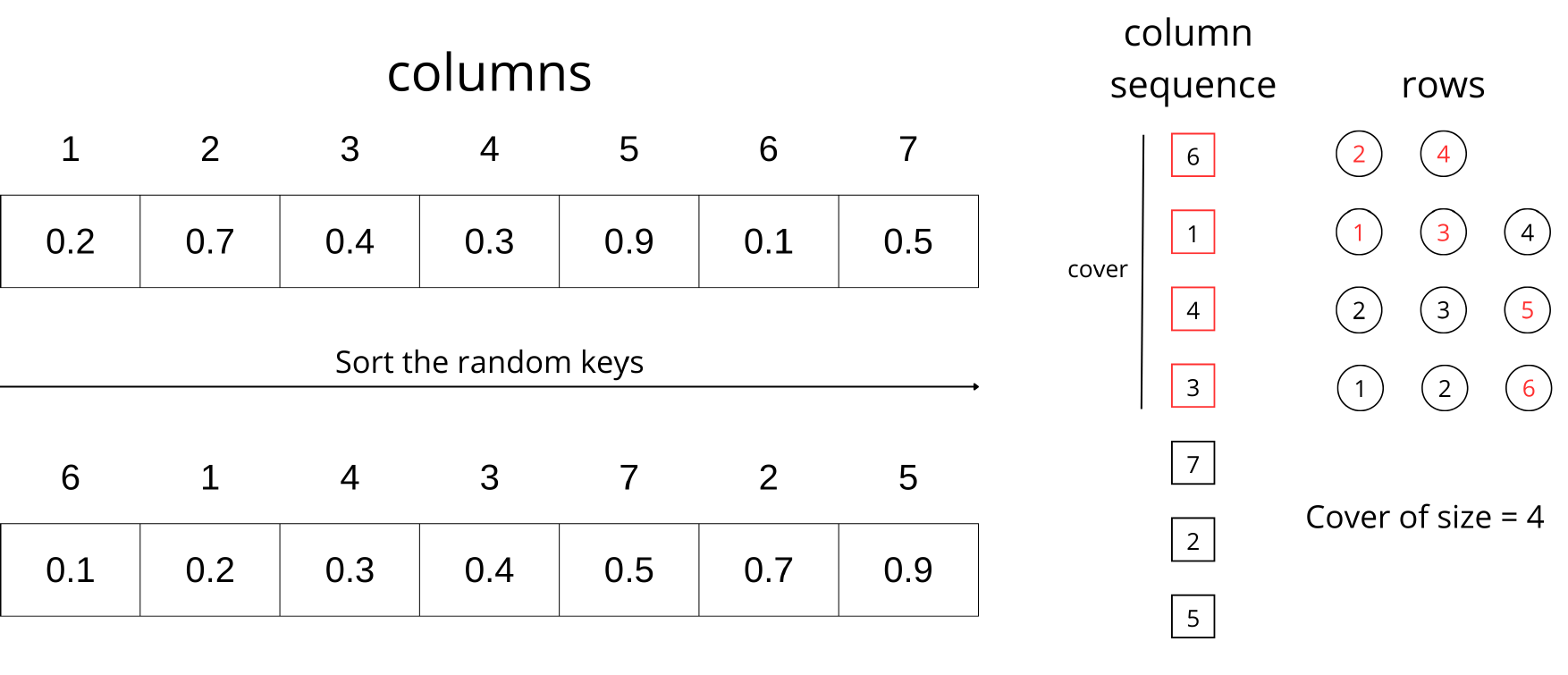}
    \caption{Example of the STCP decoder with seven elements and the incident matrix of Fano Plane.}
    \label{fig:decoderSTCP}
\end{figure}



\newpage
\subsection{Node Capacitated Graph Partitioning Problem.} 

\subsubsection{Problem definition.}
The node capacitated graph partitioning problem is also known as the handover minimization problem described as follows. Let $B$ denotes the set of base stations, with $T_b$ representing the total traffic handled by base station $b \in B$ and its connected transceivers. Let $N$ be the set of RNCs (Radio Network Controllers), where each $\mathrm{RNC}$ $r \in N$ can handle a maximum traffic capacity of $C_r$. Furthermore, define $H_{b_1, b_2}$ as the total number of handovers between base stations $b_1$ and $b_2$ ($b_1, b_2 \in B, b_1 \neq b_2$). Note that $H_{b_1, b_2}$ and $H_{b_2, b_1}$ may differ.

The objective of the handover minimization problem is to allocate each base station in $B$ to a specific RNC in $N$. Let $\rho_b$ denote the index of the RNC to which base station $b \in B$ is assigned, and let $\Psi_r$ denote the indices of base stations assigned to RNC $r \in N$. The assignments must satisfy the capacity constraints of each RNC, meaning that for all $r \in N$:

\[
\sum_{b \in \Psi_r} T_b \leq C_r.
\]

Among all feasible assignments, we seek to minimize the total handovers between base stations assigned to different RNCs:

\[
\sum_{b_1, b_2 \in B \mid \rho_{b_1} \neq \rho_{b_2}} H_{b_1, b_2}.
\]

Ferreira et al. \cite{FMSWW1998} were first to study this problem, proposing strong valid inequalities for a branch and cut algorithm. They tested their algorithm on three applications of partitioning: compiler design, finite element computations associated with meshes, and design of electronic circuits. The largest instances they were able to solve to optimality had 61 nodes and 187 edges for compiler design, 48 nodes and 81 edges for design of electronic circuits, and 274 nodes and 469 edges for finite element computations on a mesh. 
Mehrotra and Trick \cite{MEHROTRA19981} proposed a branch and price algorithm for the node capacitated graph partitioning problem. They tested their algorithm on instances having 30-61 nodes and 47-187 edges. The largest instances they were able to solve to optimality had 61 nodes and 187 edges. 
Deng and Bard \cite{DB2011} proposed a reactive GRASP with path-relinking post-processing for the node capacitated graph partitioning problem. They tested their heuristic on instances varying in size from 30 to 82 nodes and 65 to 540 edges. They used CPLEX 11 to solve to optimality instances having 30 nodes and most of the instances having 40 nodes. They showed that their heuristic was able to find solutions having the same objective function value than those found by CPLEX on most instances, but in less time than CPLEX. On larger instances, CPLEX failed to prove optimality and found worse solutions than those found by the reactive GRASP with path-relinking for all but one instance where they were tied. They also compared their reactive GRASP heuristic with the combinatorial algorithm of Mehrotra and Trick \cite{MEHROTRA19981}. Deng, Y., Bard, J.F. \cite{DB2011} showing that their heuristic matched the optimal solutions found by the combinatorial algorithm on all but one instance where it found a suboptimal solution.

\subsubsection{Solution encoding.}

Solutions of the NCGPP are encoded as a vector of $n=|B|+1$ random keys. The first $|B|$ positions represent each base station, while the last random key indicates the number of base stations assigned initially. Once sorted, the $|B|$ random keys will dictate the order in which base stations are assigned to the RNCs.

\subsubsection{Solution decoder.}


To decode a random-key vector and produce an assignment of base stations to RNCs, the following steps are computed: sort the random-key vector and then group base stations into RNCs, considering capacity constraints and costs. The assignment process starts by ensuring the first $\#N = \left \lceil  x_n \cdot |N| \right \rceil$ base stations are assigned to separate RNCs. For each subsequent base station ($|B| - \#N$), it searches for the best RNC that can accommodate the base station without exceeding capacity, evaluating the insertion cost based on the sum of handovers between the current base station and base stations that are already in the RNCs. If a suitable RNC is found, the base station is added, and the RNC’s capacity is updated. A penalty is added to the solution if no RNC can accommodate the base station. Finally, the decoder calculates the objective function value by summing the handover between base stations of different RNCs and the penalties for ungrouped base stations.

Figure \ref{fig:decoderNCGPP} shows an example of the NCGPP decoder with six base stations ($|B| = 6$) and two RNCs ($|N|=2$). In this example, $\#N = \left \lceil  0.7 \cdot 2 \right \rceil = 2$ and the ordered random-key vector provides the sequence of base stations: $2, 4, 5, 3, 6, 1$. From these $\#N$ and sequence, base stations $2$ and $4$ are allocated to separate RNCs. Base station $5$ is then allocated to RNC 2, as base station $5$ has 191 handovers with base station $4$ and only 116 with base station $2$. Base station $3$ is allocated to RNC 1, as it has no handover with base stations $4$ and $5$, only with base station $2$. Base station $6$ is allocated to RNC 2, which has 307 handovers (157 with base station $4$ and 150 with base station $5$) against 13 for RNC 1. Finally, base station $1$ is allocated to RNC 1 because it also has handovers only with base stations $2$ and $3$.

\begin{figure}[bhtp]
    \centering
    \includegraphics[width=0.75\linewidth]{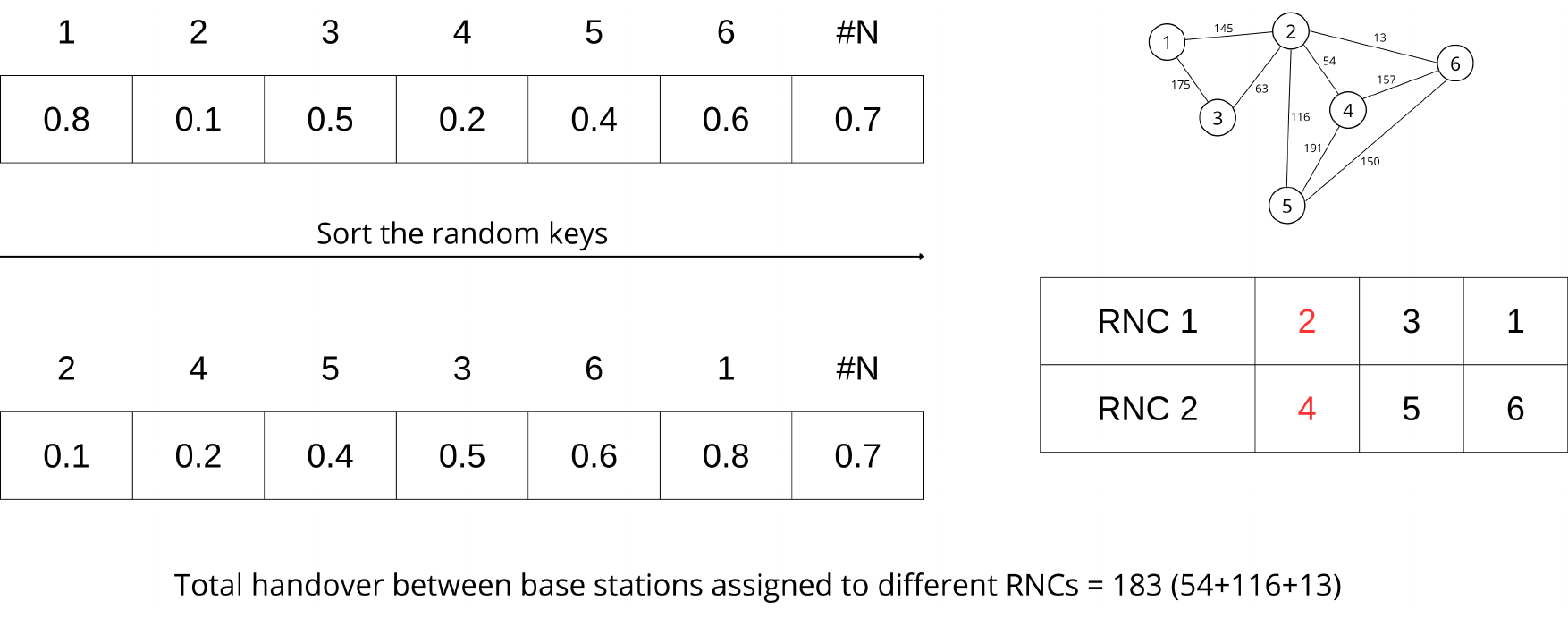}
    \caption{Example of the NCGPP decoder with six base stations and two RNCs.}
    \label{fig:decoderNCGPP}
\end{figure}



\section{Computational experiments}
\label{s_comp}

The RK-GRASP framework was coded in C++ and compiled with GCC. The computer used in all experiments was a Dual Xenon Silver 4114 20c/40t 2.2Ghz processor with 96GB of DDR4 RAM and running CentOS 8.0 x64. We tested $20$ instances of the TSP, $36$ instances of the THLP, five instances of the STCP, $20$ instances of the NCGPP, and $40$ instances of the SSP. Each algorithm was run $5$ times for each instance with different seeds and a CPU time limit ($run_{time}$). \textcolor{black}{The CPU time limit was configured in seconds, taking into account the size of each tested instance. For the TSP, the CPU time corresponds to the number of cities in the instance. For the THLP, it is based on the number of nodes in the network; for the STCP, it corresponds to the number of columns; for the SSP, it is determined by the number of tasks; and for the NCGPP, it is based on the number of base stations.} This stopping criterion was selected to allow a fairer comparison between different methods using the same architecture. However, it can be replaced by other criteria, such as the maximum number of objective function evaluations. \textcolor{black}{The RK-GRASP has three parameters to be configured: $h_s$, $h_e$, and $\theta$. These parameters are tuned based on previous computational tests as $h_s = 0.125$, $h_e=0.00098$, and $\theta =$ {\fontfamily{pcr}\selectfont UnifRand}$[0.2, 0.4]$.} The simulated annealing acceptance criterion is $e^{-\Delta/(run_{time} - time)}$, where $\Delta$ is the difference between $f(\mathcal{D}(x''))$ and $f(\mathcal{D}(x))$, and $time$ is the current CPU time. The framework, instances and detailed results are available at \url{https://github.com/antoniochaves19/RK-GRASP}.


Table~\ref{table_results} summarizes the results of the computational experiments of the three versions of the RK-GRASP applied to solve five combinatorial optimization problems. The entries in this table are the problem on hand, the instance set, the number of tested instances, the best-known solution (BKS), the average solution over five runs (Avrg), the gap between the best solution found and the BKS (Gap), the average relative percentage deviation (ARPD) considering solutions found in each run and the BKS ($RPD(\%) = \frac{(f(\mathcal{D}(x^*))-BKS)}{BKS} \times 100$), and the average running time to find the best solution. The values in boldface show the best average objective function value for each instance set.

\begin{table}[htbp]
\centering
\caption{RK-GRASP: Computational results for different combinatorial optimization problems.}\label{table_results}
\resizebox{16cm}{!}{%
\begin{tabular}{llcrrrrrrrrrrrrr} \hline
  &  & \multicolumn{1}{l}{} & \multicolumn{1}{l}{} & \multicolumn{4}{c}{RK-GRASP Grid Search} & \multicolumn{4}{c}{RK-GRASP NelderMead Search} & \multicolumn{4}{c}{RK-GRASP RVND Search} \\ \hline
Problem Class & Instance set & \# inst & BKS & Avrg & Gap & ARPD & Time (s) & Avrg & Gap & ARPD & Time (s) & Avrg & Gap & ARPD & Time (s) \\ \hline
TSP & $|N|<100$ & 5 & 27639.2 & \textbf{27639.2} & 0.00 & 0.00 & 5.09 & \textbf{27639.2} & 0.00 & 0.00 & 0.74 & \textbf{27639.2} & 0.00 & 0.00 & 0.34 \\
 & $|N|<200$ & 5 & 57468.8 & 57469.6 & 0.00 & 0.01 & 25.18 & \textbf{57468.8} & 0.00 & 0.01 & 16.43 & \textbf{57468.8} & 0.00 & 0.00 & 10.30 \\
 & $|N|<400$ & 5 & 60110.8 & 60148.6 & 0.06 & 0.34 & 104.56 & 60151.2 & 0.07 & 0.19 & 90.75 & \textbf{60121.4} & 0.02 & 0.10 & 64.48 \\
 & $|N|<1000$ & 5 & 42340.3 & 42649.5 & 0.73 & 1.03 & 311.01 & 42572.8 & 0.55 & 0.75 & 356.19 & \textbf{42398.8} & 0.14 & 0.31 & 319.95 \\
 \multicolumn{16}{l}{} \\
THLP & cab20 & 9 & 4397.9 & 4519.7 & 2.77 & 7.39 & 9.49 & 4436.2 & 0.87 & 2.84 & 8.76 & \textbf{4396.4} & -0.03 & 0.00 & 1.98 \\
 & cab25 & 9 & 7187.6 & 7540.8 & 4.91 & 9.64 & 11.26 & 7268.3 & 1.12 & 4.25 & 12.31 & \textbf{7177.7} & -0.14 & -0.06 & 4.20 \\
 & ap20 & 9 & 60397.5 & 62322.5 & 3.19 & 5.87 & 10.51 & 60928.9 & 0.88 & 2.85 & 7.02 & \textbf{60383.7} & -0.02 & -0.03 & 2.28 \\
 & ap25 & 9 & 61495.0 & 64705.5 & 5.22 & 9.31 & 13.53 & 63113.4 & 2.63 & 5.53 & 9.36 & \textbf{61443.8} & -0.08 & 0.01 & 5.20 \\
 \multicolumn{16}{l}{} \\
STCP & stn81 & 1 & 61 & \textbf{61.0} & 0.00 & 0.00 & 0.746 & \textbf{61.0} & 0.00 & 0.00 & 0.072 & \textbf{61.0} & 0.00 & 0.00 & 0.281 \\
 & stn135 & 1 & 103 & 104.0 & 0.97 & 1.17 & 20.813 & 104.0 & 0.97 & 0.97 & 51.841 & \textbf{103.0} & 0.00 & 0.78 & 1.876 \\
 & stn243 & 1 & 198 & \textbf{198.0} & 0.00 & 1.92 & 140.494 & \textbf{198.0} & 0.00 & 0.91 & 57.495 & \textbf{198.0} & 0.00 & 0.00 & 20.125 \\
 & stn405 & 1 & 335 & 344.0 & 2.69 & 2.93 & 95.396 & 343.0 & 2.39 & 2.63 & 299.486 & \textbf{339.0} & 1.19 & 1.55 & 184.676 \\
 & stn729 & 1 & 617 & 648.0 & 5.02 & 5.25 & 467.249 & 648.0 & 5.02 & 5.19 & 369.002 & \textbf{640.0} & 3.73 & 3.95 & 637.719 \\
 \multicolumn{16}{l}{} \\
NCGPP & 20/5 & 5 & 381.6 & \textbf{381.6} & 0.00 & 0.00 & 0.11 & \textbf{381.6} & 0.00 & 0.00 & 0.04 & \textbf{381.6} & 0.00 & 0.00 & 0.03 \\
 & 40/15 & 5 & 5940.0 & \textbf{5940.0} & 0.00 & 0.00 & 6.27 & \textbf{5940.0} & 0.00 & 0.01 & 1.62 & \textbf{5940.0} & 0.00 & 0.00 & 0.12 \\
 & 100/25 & 5 & 35981.2 & 36409.6 & 1.19 & 1.85 & 45.83 & 35950.8 & -0.08 & 0.19 & 45.82 & \textbf{35943.6} & -0.10 & -0.09 & 7.94 \\
 & 200/50 & 5 & 220969.6 & 220472.0 & -0.23 & 0.42 & 104.10 & 217267.6 & -1.68 & -1.34 & 165.30 & \textbf{215788.8} & -2.34 & -2.29 & 110.90 \\
 \multicolumn{16}{l}{} \\
SSP & c10-10-4 & 10 & 9.1 & \textbf{9.1} & 0.00 & 0.00 & 0.07 & \textbf{9.1} & 0.00 & 0.00 & 0.05 & \textbf{9.1} & 0.00 & 0.00 & 0.01 \\
 & c15-20-6 & 10 & 20.6 & 20.7 & 0.49 & 0.53 & 1.11 & 20.7 & 0.49 & 1.01 & 1.57 & \textbf{20.6} & 0.00 & 0.00 & 0.16 \\
 & c30-40-15 & 10 & 92.7 & 95.2 & 2.70 & 4.38 & 12.80 & 93.7 & 1.08 & 2.63 & 10.84 & \textbf{91.7} & -1.08 & -0.41 & 7.35 \\
 & c40-60-20 & 10 & 182.7 & 189.8 & 3.89 & 5.68 & 19.49 & 183.9 & 0.66 & 2.35 & 20.74 & \textbf{179.8} & -1.59 & -0.76 & 15.41 \\ \hline
\end{tabular}}
\end{table}

\textcolor{black}{The three versions of the RK-GRASP generated effective solutions for the problems on hand. The RK-GRASP variant with Grid Search found high-quality solutions for the TSP, STCP, and NCGPP, improving the Best Known Solution (BKS) of the literature for the large-scale instances of the NCGPP. The RK-GRASP with Nelder-Mead Search found better results than RK-GRASP with Grid Search, being more competitive also for the THLP and SSP in terms of quality solution. However, the RK-GRASP variant with RVND Search emerged as the most effective, delivering better solutions for the five combinatorial optimization problems. The results of this version are competitive for the TSP and STCP and for the THLP, NCGPP, and SSP this variant outperformed state-of-the-art methods, presenting several negative gaps. We can observe that the RK-GRASP converges to high-quality solutions in an efficient CPU time. In summary, all versions of RK-GRASP demonstrated efficiency in both solution quality and computational time, showing their effectiveness across diverse problems and instances.}



					
					


To assess the efficacy of incorporating the random-key concept into GRASP, we conducted experiments employing the same decoders with the BRKGA metaheuristic. Additionally, we employed a Multi-Start algorithm, in which random-key vectors are randomly generated and decoded, to highlight the significance of the RK-GRASP and BRKGA methodologies in achieving high-quality solutions. The results are presented in Tables \ref{tab:wilcoxon} and \ref{table_resultsAll}, with detailed information in the GitHub repository. \textcolor{black}{Table \ref{table_resultsAll} presents the overall results for the five combinatorial optimization problems. For each problem and algorithm, the median, the mean, and the average gap values across all runs and instances are shown. Additionally, the table includes the number of instances tested for each problem, the number of instances where the algorithm matched the best-known solution from the literature (\#BKS), and the number of instances where the algorithm found a new best-known solution (\#newBKS).} We also conducted a Wilcoxon signed-rank test (WSR) \cite{siegal1956nonparametric}, a non-parametric statistical hypothesis test, to compare the rank sums between each pair of methods (Table \ref{tab:wilcoxon}). In interpretation, a $p$-value $< 0.05$ with alternative = ``less'' implies that the first group tends to produce smaller values than the second. In contrast, a $p$-value $< 0.05$ with alternative = ``greater'' indicates that the first group tends to produce larger values. \textcolor{black}{The values in boldface show the cases with statistical significance ($p$-value $< 0.05$).}

\begin{table}[htbp]
\centering
\caption{Wilcoxon test comparing the different algorithms in terms of $p$-value.}\label{tab:wilcoxon}
\resizebox{\textwidth}{!}{%
\begin{tabular}{lrrrrr} \hline
\multicolumn{6}{c}{Alternate hypothesis: solution set A is less than solution set B (A x B)} \\ \hline
Problem Class & TSP & THLP & STCP & NCGPP & SSP \\  \hline
RK-GRASP Grid Search x RK-GRASP NelderMead Search & 9.99E-01 & 1.00E+00 & 9.96E-01 & 1.00E+00 & 1.00E+00 \\
RK-GRASP Grid Search x RK-GRASP RVND Search & 1.00E+00 & 1.00E+00 & 1.00E+00 & 1.00E+00 & 1.00E+00 \\
RK-GRASP Grid Search x BRKGA & 1.00E+00 & 1.00E+00 & 9.98E-01 & 1.00E+00 & 1.00E+00 \\
RK-GRASP Grid Search x MultiStart & \textbf{8.46E-14} & \textbf{2.20E-16} & 1.22E-01 & \textbf{8.46E-14} & \textbf{2.20E-16} \\
RK-GRASP NelderMead Search x RK-GRASP RVND Search & 1.00E+00 & 1.00E+00 & 9.99E-01 & 1.00E+00 & 1.00E+00 \\
RK-GRASP NelderMead Search x BRKGA & 1.00E+00 & 1.00E+00 & 9.86E-01 & 1.00E+00 & 1.00E+00 \\
RK-GRASP NelderMead Search x MultiStart & \textbf{8.46E-14} & \textbf{2.20E-16} & \textbf{3.60E-03} & \textbf{8.46E-14} & \textbf{2.20E-16} \\
RK-GRASP RVND Search x BRKGA & \textbf{3.03E-02} & \textbf{2.20E-16} & \textbf{1.42E-02} & \textbf{2.48E-16} & \textbf{3.91E-09} \\
RK-GRASP RVND Search x MultiStart & \textbf{8.46E-14} & \textbf{2.20E-16} & \textbf{3.15E-04} & \textbf{8.46E-14} & \textbf{2.20E-16} \\
 & \multicolumn{1}{l}{} & \multicolumn{1}{l}{} & \multicolumn{1}{l}{} & \multicolumn{1}{l}{} & \multicolumn{1}{l}{} \\ \hline
\multicolumn{6}{c}{Alternate hypothesis: solution set A is greater than solution set B (A x B)} \\ \hline
Problem Class & TSP & THLP & STCP & NCGPP & SSP \\ \hline
RK-GRASP Grid Search x RK-GRASP NelderMead Search & \textbf{9.24E-04} & \textbf{2.20E-16} & \textbf{6.44E-03} & \textbf{3.90E-10} & \textbf{8.75E-16} \\
RK-GRASP Grid Search x RK-GRASP RVND Search & \textbf{4.66E-08} & \textbf{2.20E-16} & \textbf{2.22E-04} & \textbf{3.90E-10} & \textbf{2.20E-16} \\
RK-GRASP Grid Search x BRKGA & \textbf{1.01E-07} & \textbf{2.20E-16} & \textbf{2.21E-03} & \textbf{3.90E-10} & \textbf{2.20E-16} \\
RK-GRASP Grid Search x MultiStart & 1.00E+00 & 1.00E+00 & 8.98E-01 & 1.00E+00 & 1.00E+00 \\
RK-GRASP NelderMead Search x RK-GRASP RVND Search & \textbf{5.34E-06} & \textbf{2.20E-16} & \textbf{7.62E-04} & \textbf{1.85E-08} & \textbf{2.20E-16} \\
RK-GRASP NelderMead Search x BRKGA & \textbf{9.96E-06} & \textbf{4.16E-04} & \textbf{1.78E-02} & \textbf{3.34E-06} & \textbf{2.91E-13} \\
RK-GRASP NelderMead Search x MultiStart & 1.00E+00 & 1.00E+00 & 9.97E-01 & 1.00E+00 & 1.00E+00 \\
RK-GRASP RVND Search x BRKGA & 9.71E-01 & 1.00E+00 & 9.90E-01 & 1.00E+00 & 1.00E+00 \\
RK-GRASP RVND Search x MultiStart & 1.00E+00 & 1.00E+00 & 1.00E+00 & 1.00E+00 & 1.00E+00 \\ \hline
\end{tabular}
}
\end{table}

\begin{table}[htbp]
\caption{\textcolor{black}{Comparison of computational results for the three variants of RK-GRASP, BRKGA, and Multi-Start.}}\label{table_resultsAll}
\resizebox{0.8\textwidth}{!}{%
\begin{tabular}{llrrrrr} \hline
 &  & \multicolumn{1}{c}{RK-GRASP} & \multicolumn{1}{c}{RK-GRASP} & \multicolumn{1}{c}{RK-GRASP} & \multicolumn{1}{c}{BRKGA} & \multicolumn{1}{c}{Multi-Start} \\ 
  &  & \multicolumn{1}{c}{Grid Search} & \multicolumn{1}{c}{NelderMead Search} & \multicolumn{1}{c}{RVND Search} & \multicolumn{1}{c}{} & \multicolumn{1}{c}{} \\\hline
TSP & median & 38974.5 & 38872.5 & \textbf{38694.0} & 38727.0 & 39545.0 \\
 & mean & 45248.0 & 45214.8 & \textbf{45147.6} & 45166.5 & 45810.9 \\
 & avrg gap & 0.25 & 0.24 & \textbf{0.07} & 0.12 & 1.66\\
 & \#inst & 20 & 20 & 20 & 20 & 20 \\
 & \#BKS & 13 & 14 & 14 & 14 & 6 \\
 & \#newBKS & 0 & 0 & 0 & 0 & 0 \\ 
 &  & \multicolumn{1}{l}{} & \multicolumn{1}{l}{} & \multicolumn{1}{l}{} & \multicolumn{1}{l}{} & \multicolumn{1}{l}{} \\
THLP & median & 24811.1 & 22965.5 & \textbf{22610.4} & 22962.4 & 31175.4 \\
 & mean & 35575.4 & 34583.1 & \textbf{33363.1} & 34255.5 & 42441.3 \\
 & avrg gap & 4.94 & 1.51 & \textbf{-0.06} & 1.39 & 28.72 \\
 & \#inst & 36 & 36 & 36 & 36 & 36 \\
 & \#BKS & 7 & 12 & 36 & 10 & 0 \\
 & \#newBKS & 0 & 0 & 5 & 1 & 0 \\ 
 &  & \multicolumn{1}{l}{} & \multicolumn{1}{l}{} & \multicolumn{1}{l}{} & \multicolumn{1}{l}{} & \multicolumn{1}{l}{} \\
STCP & median & 203.0 & \textbf{198.0} & \textbf{198.0} & \textbf{198.0} & 202.0 \\
 & mean & 272.2 & 271.5 & \textbf{268.8} & 270.7 & 272.4 \\
 & avrg gap & 1.74 & 1.68 & \textbf{0.98} & 1.40 & 1.83\\
 & \#inst & 5 & 5 & 5 & 5 & 5 \\
 & \#BKS & 2 & 2 & 3 & 3 & 2 \\
 & \#newBKS & 0 & 0 & 0 & 0 & 0 \\ 
 &  & \multicolumn{1}{l}{} & \multicolumn{1}{l}{} & \multicolumn{1}{l}{} & \multicolumn{1}{l}{} & \multicolumn{1}{l}{} \\
NCGPP & median & 20699.0 & \textbf{20666.0} & \textbf{20666.0} & \textbf{20666.0} & 22756.0 \\
 & mean & 66228.0 & 65109.5 & \textbf{64552.6} & 64848.4 & 70850.2 \\
 & avrg gap & 0.23 & -0.44 & \textbf{-0.62} & -0.53 & 4.62\\
 & \#inst & 20 & 20 & 20 & 20 & 20 \\
 & \#BKS & 14 & 19 & 20 & 20 & 7 \\
 & \#newBKS & 0 & 1 & 6 & 1 & 0 \\ 
 &  & \multicolumn{1}{l}{} & \multicolumn{1}{l}{} & \multicolumn{1}{l}{} & \multicolumn{1}{l}{} & \multicolumn{1}{l}{} \\
SSP & median & 52.0 & 52.5 & \textbf{51.0} & \textbf{51.0} & 64.0 \\
 & mean & 79.9 & 78.0 & \textbf{75.8} & 76.7 & 96.4 \\
 & avrg gap & 1.75 & 0.57 & \textbf{-0.67} & -0.33 & 14.95 \\
 & \#inst & 40 & 40 & 40 & 40 & 40 \\
 & \#BKS & 19 & 24 & 39 & 37 & 11 \\
 & \#newBKS & 0 & 1 & 14 & 7 & 0 \\ \hline
\end{tabular}%
}
\end{table}

\textcolor{black}{We observe that RK-GRASP with RVND emerged as the most efficient method for the five combinatorial optimization problems, with statistical differences evident among the other two variants of RK-GRASP, BRKGA, and Multi-Start. Additionally, BRKGA demonstrated statistical superiority over RK-GRASP with Grid Search and Nelder-Mead, becoming the second best choice. The RK-GRASP with Nelder-Mead performed better than RK-GRASP with Grid Search. Overall, all metaheuristics outperformed the Multi-Start approach. The RK-GRASP with RVND improved the BKS in 5 instances of THLP (14\%), 6 instances of NCGPP (30\%), and 14 instances of SSP (35\%). This RK-GRASP version also achieved a solution equal to or better than the BKS in 92.5\% of the tested instances.}


{\textcolor{black}{
The performance profile \citep{dolan2002benchmarking} was used to evaluate and compare the performance of three variants of RK-GRASP, BRKGA, and Multi-Start methods based on CPU time. The performance ratio is calculated by:
\[
r_{i,a} = \frac{t_{i,a}}{min\{t_{i,a}: \forall a\}}
\]
where $t_{i,a}$ is the average CPU time required to find the best solution to instance $i$ using algorithm $a$. We define the level of accuracy as 1\%. Each instance where an algorithm had a gap greater than 1\% was assigned an infinite CPU time ($\mathit{INF}$) to indicate failure in meeting the convergence test. 
}

{\textcolor{black}{
The cumulative distribution function $\rho_a(\tau)$ represents the probability that algorithm $a$ achieves a performance ratio $r_{i,a}$ within a factor $\tau$ of the best possible ratio:
\[
\rho_a(\tau) = \frac{\left| \{i \in I: r_{i,a} \leq \tau\} \right |}{|I|}. 
\]
Algorithms with higher $\rho_a(\tau)$ values are preferred.
}

Figure \ref{fig:perfProfile} presents a $log_2$-scaled view of the performance profiles of the three variants of RK-GRASP, BRKGA, and Multi-Start methods. \textcolor{black}{Again, we observe the dominance of the RK-GRASP with RVND in terms of CPU time. This variant of the RK-GRASP was the most efficient method in 97\% of THLP instances, 50\% of NCGPP instances, and 83\% of SSP instances, while BRKGA was the most efficient method in 60\% of TSP instances. These two methods always dominate the other methods, except in the case of STCP, where no visible dominance is evident. Considering a performance rate of $\tau = 4$ (x-axis = 2), we note that BRKGA can find the target solution for 95\% of TSP, 15\% of THLP, 40\% of STCP, 80\% of NCGPP, and 73\% of SSP instances, while RK-GRASP RVND solved 95\% of TSP, 100\% of THLP, 40\% of STCP, 75\% of NCGPP, and 100\% of SSP instances.}


\begin{figure}[htpb]
     \centering
     \scriptsize
     \begin{subfigure}[b]{0.45\textwidth}
         \centering
         \includegraphics[width=\textwidth]{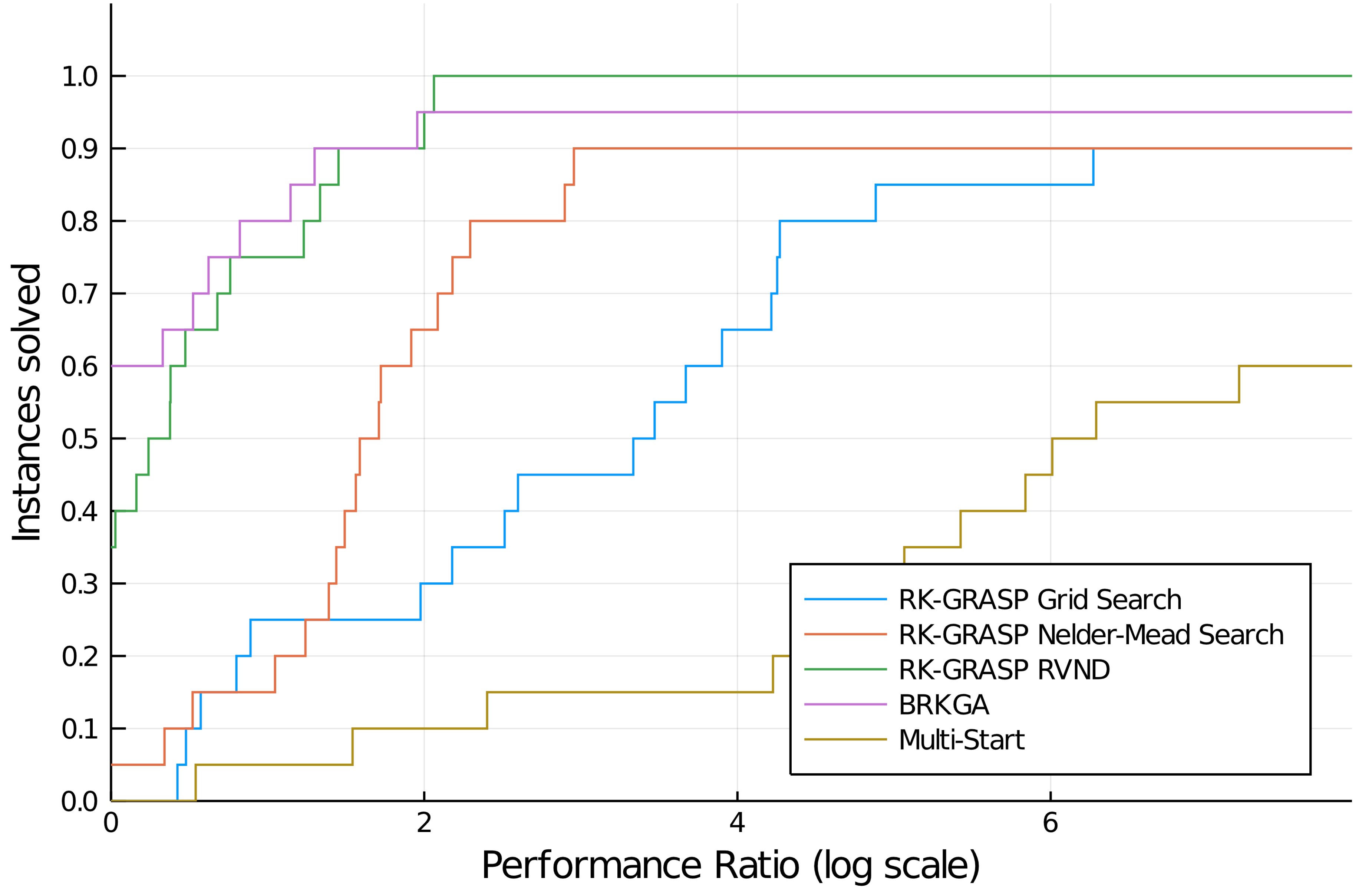}
         \caption{TSP.}
         \label{fig:pp_1}
     \end{subfigure}
     \hfill
     \begin{subfigure}[b]{0.45\textwidth}
         \centering
         \includegraphics[width=\textwidth]{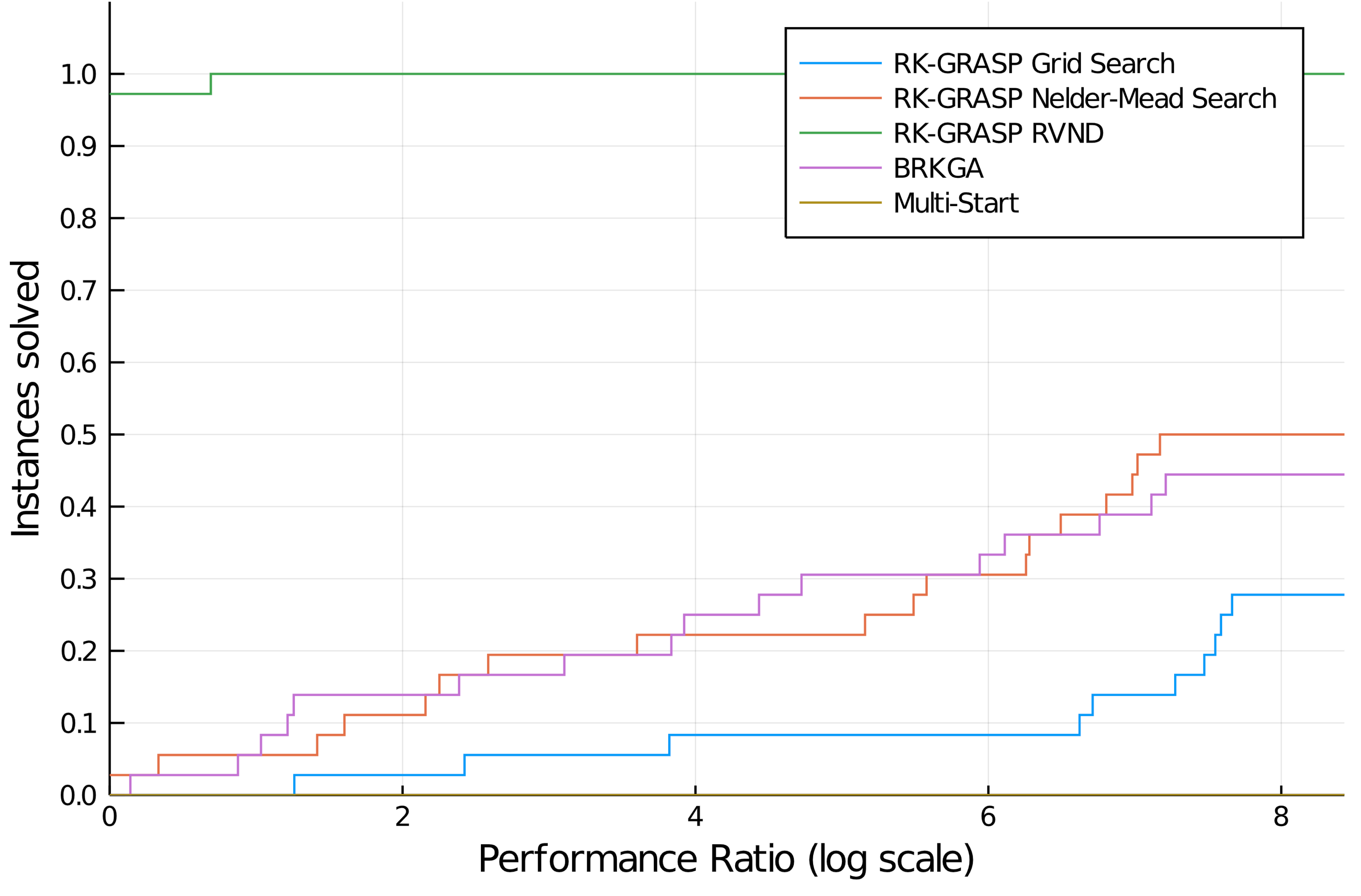}
         \caption{THLP.}
         \label{fig:pp_2}
     \end{subfigure}
     \hfill
     \begin{subfigure}[b]{0.45\textwidth}
         \centering
         \includegraphics[width=\textwidth]{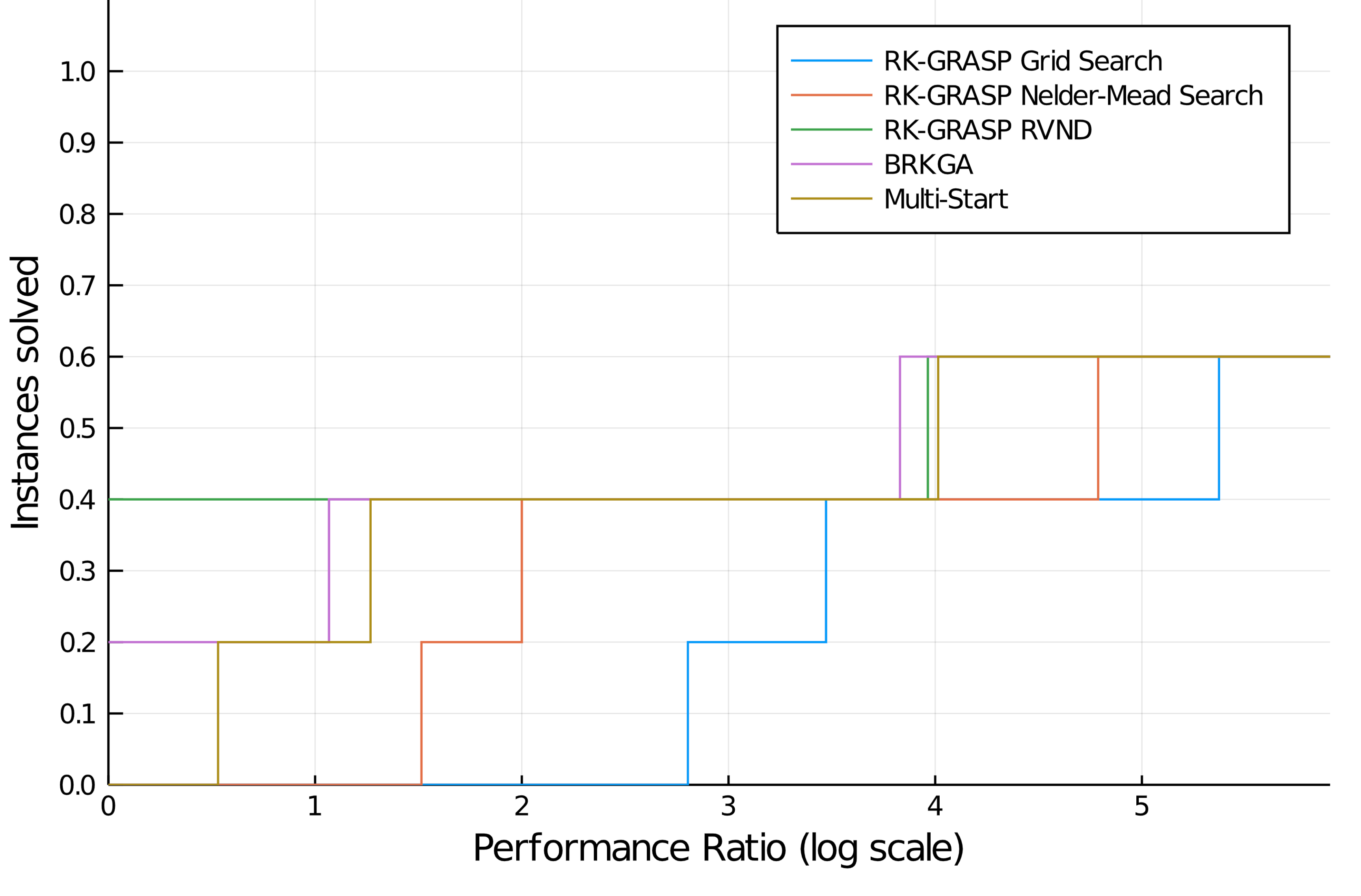}
         \caption{STCP.}
         \label{fig:pp_3}
     \end{subfigure}
    \hfill
     \begin{subfigure}[b]{0.45\textwidth}
         \centering
         \includegraphics[width=\textwidth]{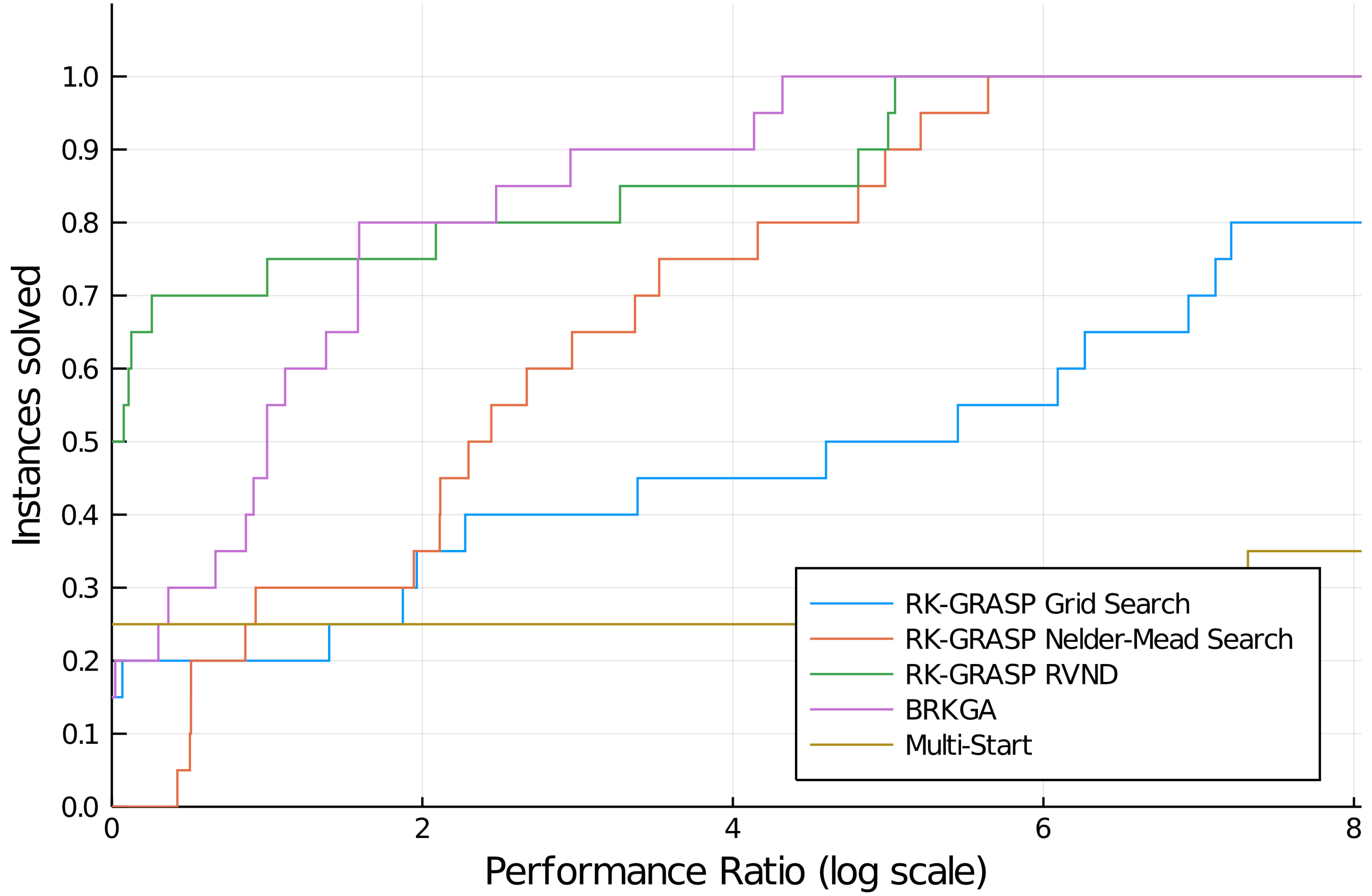}
         \caption{NCGPP.}
         \label{fig:pp_4}
     \end{subfigure}
     \hfill
          \begin{subfigure}[b]{0.4\textwidth}
         \centering
         \includegraphics[width=\textwidth]{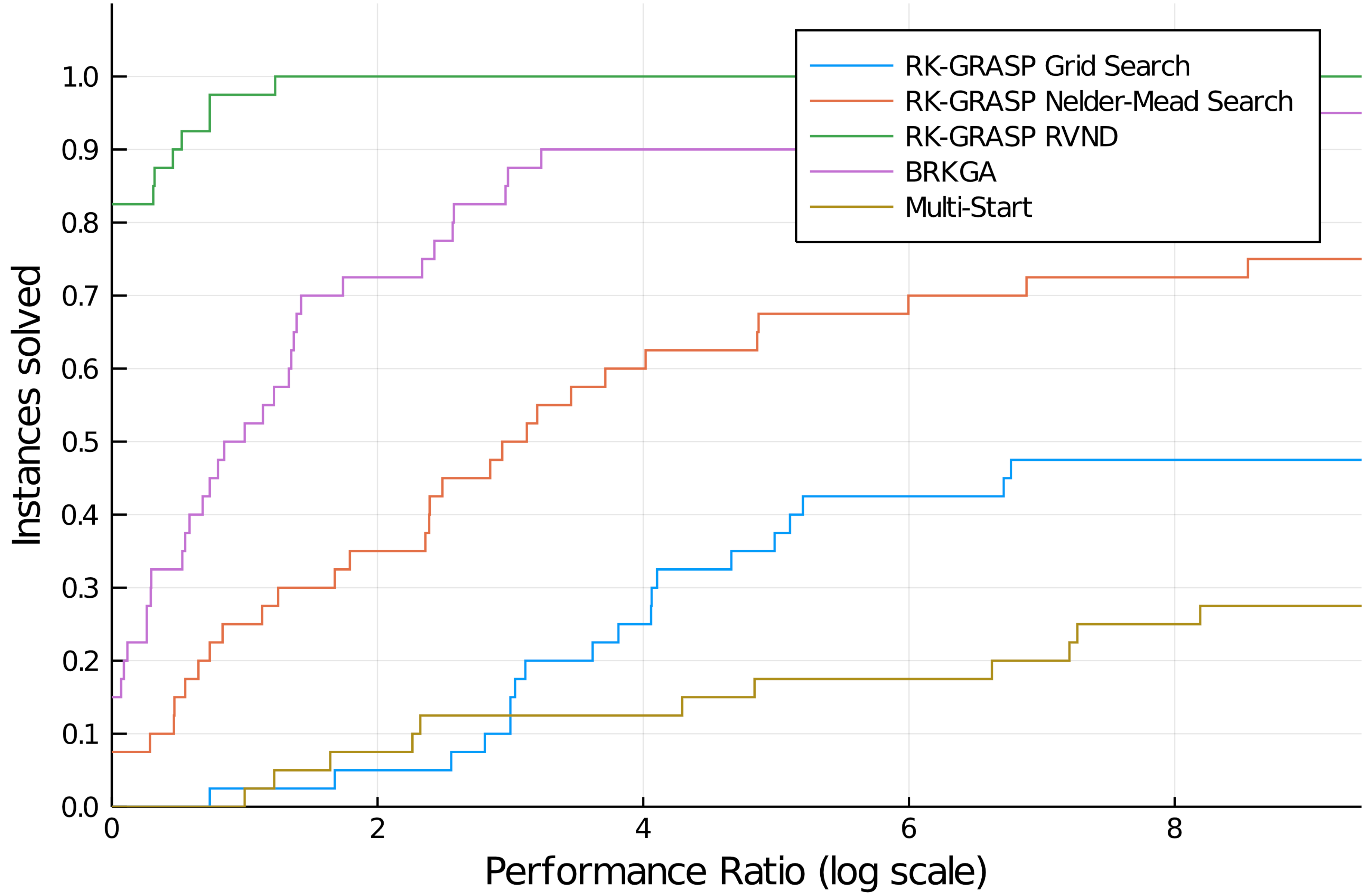}
         \caption{SSP.}
         \label{fig:pp_5}
     \end{subfigure}
        \caption{Performance profile of runtime for the RK-GRASP versions to find the best solution.}
        \label{fig:perfProfile}
\end{figure}

\section{Concluding remarks}
\label{s_concl}

This paper introduces \textit{RK-GRASP}, a problem-independent greedy randomized adaptive search procedure (GRASP) for combinatorial optimization, leveraging the random-key optimizer (RKO) paradigm. RK-GRASP distinguishes itself from standard GRASP by operating in the continuous unit hypercube and employing a decoder to map solutions to the discrete problem space. This design allows for a problem-independent implementation, requiring only the development of a specific decoder for each problem.

Three local search methods, namely Grid Search, Nelder-Mead Search, and Random Variable Neighborhood Descent (RVND), are integrated into RK-GRASP.  The paper demonstrates the applicability of RK-GRASP across five NP-hard combinatorial optimization problems: Traveling Salesman Problem (TSP), Tree of Hubs Location Problem (THLP), Steiner Triple Covering Problem (STCP), Node Capacitated Graph Partitioning Problem (NCGPP), and Job Sequencing and Tool Switching Problem (SSP).  

\textcolor{black}{Computational experiments show that all three versions of RK-GRASP generate effective solutions. RK-GRASP with Nelder-Mead Search performs well for TSP and NCGPP instances, while RK-GRASP with RVND Search emerges as the most effective for the five combinatorial optimization problems. Notably, RK-GRASP with RVND Search outperforms state-of-the-art methods for THLP, SSP, and NCGPP, achieving new best-known solutions.}

\textcolor{black}{The paper also compares RK-GRASP with BRKGA and a Multi-Start approach using the same decoders. Again, RK-GRASP with RVND demonstrates significant statistical superiority and is more effective for the problems on hand.}

In conclusion, this paper presents RK-GRASP as a versatile and efficient problem-independent metaheuristic for combinatorial optimization. \textcolor{black}{The RVND was the best choice of local search technique within RK-GRASP, considering the specific problems and tested instances.} The results highlight the effectiveness of the random-key concept within the GRASP framework, offering a promising approach for tackling complex optimization challenges across diverse problem domains. 

As we did for BRKGA in \cite{AndTosGonRes21a, OliCarOliRes22a, TosRes15a}, we envision a future where an API for RK-GRASP will facilitate the use of this metaheuristic to solve a wide range of combinatorial optimization problems.


\vskip 6mm
\noindent{\bf Acknowledgments}

\noindent Ant\^onio A. Chaves was supported by S\~ao Paulo Research Foundation (FAPESP) under grants 2018/15417-8 and 2022/05803-3, and Conselho Nacional de Desenvolvimento Científico e Tecnológico (CNPq) under grants 312747/2021-7 and 405702/2021-3. Ricardo M. A. Silva was partially supported by the Foundation for the Support of Development of the Federal University of Pernambuco, Brazil (FADE), the Office for Research and Graduate Studies of the Federal University of Pernambuco (PROPESQI),and the Foundation for Support of Science and Technology of the State of Pernambuco (FACEPE).

\bibliographystyle{plainnat}
\bibliography{References}


\end{document}